\documentclass{article}

\usepackage{arxiv}

\usepackage[utf8]{inputenc} 
\usepackage[T1]{fontenc}    
\usepackage{hyperref}       
\usepackage{url}            
\usepackage{booktabs}       
\usepackage{amsfonts}       
\usepackage{nicefrac}       
\usepackage{microtype}      
\usepackage{lipsum}		
\usepackage{graphicx}
\usepackage{natbib}
\usepackage{doi}
\usepackage{verbatim}
\usepackage{array} 
\usepackage{longtable}
\usepackage{geometry}

\title{Large Language Model Agent for Structural Drawing Generation Using ReAct Prompt Engineering and Retrieval Augmented Generation}

\author{ \href{https://orcid.org/0000-0001-7615-8013}{\includegraphics[scale=0.06]{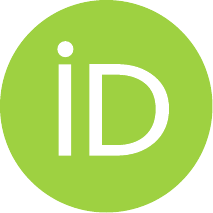}\hspace{1mm}Xin Zhang}\thanks{Corresponding author: Xin Zhang. Email: xinzhang@alumni.purdue.edu} \\
	Lyles School of Civil and Construction Engineering\\
	Purdue University\\
	West Lafayette, IN 47906 \\
	\texttt{zhan3794@purdue.edu} \\
	\And
	\href{}{Lissette Iturburu} \\
	Lyles School of Civil and Construction Engineering\\
	Purdue University\\
	West Lafayette, IN 47906 \\
	\texttt{liturbur@purdue.edu} \\
    \And
	\href{}{Juan Nicolas Villamizar} \\
	Lyles School of Civil and Construction Engineering\\
	Purdue University\\
	West Lafayette, IN 47906 \\
	\texttt{jvillam@purdue.edu} \\
    \And
	\href{}{Xiaoyu Liu} \\
	School of Mechanical Engineering\\
	Purdue University\\
	West Lafayette, IN 47906 \\
	\texttt{liu1787@purdue.edu} \\
    \And
	\href{}{Manuel Salmeron} \\
	Lyles School of Civil and Construction Engineering\\
	Purdue University\\
	West Lafayette, IN 47906 \\
	\texttt{salmeron@purdue.edu} \\
    \And
	\href{}{Shirley J. Dyke} \\
	Lyles School of Civil and Construction Engineering\\
    School of Mechanical Engineering \\
	Purdue University\\
	West Lafayette, IN 47906 \\
	\texttt{sdyke@purdue.edu} \\
    \And
	\href{}{Julio Ramirez} \\
	Lyles School of Civil and Construction Engineering\\
	Purdue University\\
	West Lafayette, IN 47906 \\
	\texttt{ramirez@purdue.edu} \\
}

\renewcommand{\shorttitle}{\textit{arXiv}}

\hypersetup{
pdftitle={A template for the arxiv style},
pdfsubject={q-bio.NC, q-bio.QM},
pdfauthor={Xin Zhang},
pdfkeywords={Structural Drawing, Generative Artificial Intelligence, Large Language Model, Transformer, Prompt Engineering, ReAct, Retrieval Augmented Generation},
}

\begin{document}
\maketitle
\begin{abstract}
Structural drawings are widely used in many fields, e.g., mechanical engineering, civil engineering, etc. In civil engineering, structural drawings serve as the main communication tool between architects, engineers, and builders to avoid conflicts, act as legal documentation, and provide a reference for future maintenance or evaluation needs. They are often organized using key elements such as title/subtitle blocks, scales, plan views, elevation view, sections, and detailed sections, which are annotated with standardized symbols and line types for interpretation by engineers and contractors. Despite advances in software capabilities, the task of generating a structural drawing remains labor-intensive and time-consuming for structural engineers. Here we introduce a novel generative AI-based method for generating structural drawings employing a large language model (LLM) agent. The method incorporates a retrieval-augmented generation (RAG) technique using externally-sourced facts to enhance the accuracy and reliability of the language model. This method is capable of understanding varied natural language descriptions, processing these to extract necessary information, and generating code to produce the desired structural drawing in AutoCAD. The approach developed, demonstrated and evaluated herein enables the efficient and direct conversion of a structural drawing's natural language description into an AutoCAD drawing, significantly reducing the workload compared to current working process associated with manual drawing production, facilitating the typical iterative process of engineers for expressing design ideas in a simplified way.
\end{abstract}


\section{Introduction}
Structural drawings contain essential information for the construction of buildings, bridges, stadiums or any other type of structure [\cite{brito2016drawing}]. Ensuring compliance through these drawings is fundamental for the integrity, safety, and functionality of those structures during their life cycle. Structural drawings typically form a complete package when integrated with drawings contributed from other disciplines, to be presented as part of the construction permit for a given project. Therefore, drawing sets must comply with regulatory standards and be signed by licensed professional engineers who carry liability.   

Historically, structural drawings were generated entirely by hand. This work required a high-level of skill and training. Draftsmen used pencils, pens, rulers, compasses, and protractors to create detailed representations on paper. The introduction of drawing reproduction techniques during the Industrial Revolution, such as lithography and engraving, marked a significant change. These methods allowed for multiple copies of a drawing to be produced, greatly enhancing the sharing and dissemination of complex designs. The leap to digital methods began with the introduction of computer graphics and Computer Aided Design (CAD) in the latter half of the 20th century [\cite{connolly1999cad},\cite{buhr1989software}]. This transition revolutionized technical drawing by greatly improving the efficiency and accuracy of the drawings. CAD software allowed for the creation of detailed 2D and 3D models with greater precision and ease of modification compared to manual drafting. For example, AutoCAD [\cite{zakaria2012effectiveness}], well-known software developed by Autodesk in 1982, has been widely used to generate structural and mechanical drawings in many fields. The integration of CAD into engineering and design processes signified a major shift, connecting the engineering analysis packages, such as SAP2000, CSI Bridge, to these drafting tools, making design processes faster, more precise, and more flexible [4]. Later developments, such as Building Information Modeling (BIM) employed software such as REVIT to build upon CAD alone. BIM-based software not only creates a 3D model but also enables a workflow to support linking and management of lifecycle building data, from the earliest planning stages to its eventual demolition [\cite{volk2014building}]. 

The transition from traditional drafting to digital modeling marks a significant evolution in this field, enhancing efficiency. However, several challenges remain inherent in this laborious and critical task. These challenges include: (1) \textbf{Time}: despite the advances made possible with software like AutoCAD, generating structural drawings is still a time-intensive task. Drawings require details and specifications to represent real-life situations that cannot be easily represented using the AutoCAD capabilities given the nature of the problem, taking more time than desired. (2) \textbf{Training}: for structural drawings, individuals need to have experience in the specific domain. Such knowledge includes an understanding of civil engineering principles, structural engineering concepts, the accepted norms of structural drawings, and proficiency in AutoCAD software itself. Acquiring and integrating this knowledge is a significant endeavor, demanding both formal education and practical experience in these domains [\cite{tavakoli1991construction}]. Structural drawings can have high variability due to different structure types like buildings, bridges, etc., and different materials like concrete, steel, etc. And each structure type and material have different specifications that are not easy to transfer. (3) \textbf{Potential for Inconsistencies}: a major challenge in the creation of structural drawings is ensuring consistency in the description and representation of various elements. Inconsistencies in these drawings can lead to errors in the final output. These errors might stem from discrepancies in interpreting the design, miscommunication among team members, or simple human error during the drafting process. Such inaccuracies can have serious implications for the structural integrity and safety of the resulting construction. Moreover, this process becomes repetitive for each project in any company as different people participate, leading to the same situations that can be often avoided if learning process from this type of situations can be reused more easily. (4) \textbf{Secondary Steps in the Design Process}: The drawing generation process often requires secondary steps to get the details as needed. Verifying conflicts, checking cover thicknesses, reinforcement positions, overlaps are just a few examples where drafters must do manual checks where mistakes can be made using AutoCAD tools. Such tasks are not complex but with a certain amount of minor calculation, the time spent on the structure generation work and the possibility of errors in the results could increase.

The adoption of artificial intelligence (AI) techniques has significant promise to address these challenges. In recent years, AI has been increasingly utilized to assist various work in the civil engineering field. A prominent application is image classification through deep neural networks. Kim et al. [\cite{kim2019crack}] introduced a classification framework to identify cracks amidst crack-like patterns using CNN and Speeded-Up Robust Features (SURF). Yeum et al. [\cite{yeum2017rapid}] employed AlexNet for efficient documentation of post-event building images, leading to the development of a powerful web-based tool for field application. Dang et al. [\cite{dang2021mixed}] pioneered the use of UAVs for capturing close-up images of bridges, followed by the automatic detection of structural damage using CNNs on image patches. The use of technologies like object detection and semantic segmentation has also been prominent. Hoskere et al. [\cite{hoskere2018vision}] applied multiscale deep convolutional neural networks for pixel-wise classification of images at multiple scales, enabling the recognition of various types of damage. Cha et al. [\cite{cha2017deep}] explored the application of Faster R-CNN, a region-based method by Ren et al. [\cite{ren2015faster}], for the rapid identification of multiple damages, including concrete cracks, corrosion levels, and delamination. Zhang et al. [\cite{zhang2018deep}] introduced CrackNet, an efficient architecture for the semantic segmentation of pavement cracks. Iturburu et al. [\cite{iturburu2023towards}] utilized semantic segmentation techniques for identifying columns and walls areas in structural drawings, aiding in assessing the vulnerability of concrete buildings. Additionally, reinforcement learning techniques have been implemented to assist in inspection and maintenance management. Wei et al. [\cite{wei2020optimal}] devised a reinforcement learning-based framework for determining the optimal policy for structural maintenance. Furthermore, Zhang et al. [\cite{zhang2023reinforcement}] employed deep Q learning, a reinforcement learning algorithm, for optimizing the timing and methods of bridge inspections. 

The introduction of generative AI (GenAI) techniques into civil engineering will further offer opportunities for practicing engineers. Generative AI is widely used for data augmentation. Luleci et al. [\cite{luleci2023generative}] utilized generative adversarial networks for labeled acceleration data augmentation for structural damage detection. Marano et al. [\cite{marano2024generative}] also introduced Generative Adversarial Networks (GAN) is adopted for generating synthetic seismic signals. For damage identification, Rastin et al. [\cite{rastin2021generative}] proposed a novel two-stage technique based on generative adversarial networks for unsupervised structural health monitory and damage identification. Shoemaker et al. [\cite{shoemaker2023generative}] discussed the possibility of using generative AI technology to develop a geotechnical assistant as LLMs like ChatGPT-4 performed well on academic and professional exams. However, the author also indicated the necessity of monitoring the quality control due to the uncertainty in the quality of the content generated by generative AI models. 

For structural design and construction, Feng et al. [\cite{feng2023intelligent}] developed a rule-embedded GAN called StructGAN-Rule to address the demand for a rapid and accurate cross-sectional design of shear wall components. Ghimire et al. [\cite{ghimire2024opportunities}] discussed the opportunities and challenges of Generative AI in Construction Industry, particularly for the adoption of text-based models. In the work, a conceptual GenAI implementation framework is recommended. Recently, the Transformer architecture [\cite{ashish2017attention}], with its attention mechanism [\cite{vasan2024detection}], has initiated a new revolution in AI, leading to innovative and more reliable applications. One such advancement includes using Transformer models for detecting surface damage through images [\cite{zhao2023survey}], showcasing its efficacy in visual data analysis. Furthermore, with the widespread adoption of generative AI (GenAI) models like LLM [\cite{zhao2023survey}], built on Transformer architecture, has potential to open up novel avenues in civil engineering applications. LLMs are advanced artificial intelligence systems designed to understand and generate human-like text by processing vast amounts of data. These models, which include examples like OpenAI's ChatGPT (Generative Pre-trained Transformer, [\cite{roumeliotis2023chatgpt}]), can handle complex language tasks, from translation and summarization to question answering and conversation. Besides, many GenAI companies like OpenAI and Anthropic also use LLM as an agent to control the manipulation of a computer or machine. These LLM techniques can offer capabilities of assisting in the civil engineering field. For example, Martin Aluga [\cite{aluga2023application}] discussed the application of LLMs, particularly ChatGPT, in design and planning, structural analysis and simulation, code compliance and regulations, construction management, knowledge repository and information retrieval, etc. Another example of applications is given by Rane et al. [\cite{rane2023integrating}] to integrate Building Information Modeling (BIM) with ChatGPT and other similar generative artificial intelligence, for optimizing efficiency of using texted BIM information.

Considering the challenges associated with generating structural drawings and the opportunities brought by new AI techniques, this paper proposes an LLM-assisted structural drawing generation method. The objective of this method is to directly transform an engineer’s natural language description of a structural drawing to  Python code which can produce corresponding structural drawing in AutoCAD software. In the method, the LLM takes the natural language description as the input and then analyzes the input for generating the Python code. The LLM considered has two common limitations for this task. The first limitation is the number of tokens (referring to the smallest unit of text that the model can process, which may be a part of a word, a whole word, or even punctuation). The second limitation is that the task to be completed in this work is quite complex for an LLM. Therefore, for our work, we build a pipeline including several LLMs. Each LLM in the pipeline conducts a specific subtask so the size of the generation and the complexity of the subtask for each LLM can be reduced. This LLM design can improve the efficiency and reliability of the generation. Due to the limitations of data and computational source for pretraining or finetuning, prompt engineering [\cite{giray2023prompt}] is adopted for instructing the LLMs to complete the structural drawing generation task. The prompt with ReAct framework [\cite{yao2023react}], which is a manner requesting LLMs to generate both reasoning traces and task-specific actions to improve the model’s ability, is established for each LLM in the pipeline. Additionally, to reduce the potential for hallucination in the LLM [\cite{yao2023llm}], a retrieval augmented generation (RAG) [\cite{lewis2020retrieval}] technique is introduced and necessary background information regarding the requested structural drawing is provided to the LLMs in the pipeline as the external source of data. This technique provides supplemental information which can control the LLM to follow the right path. Engineers can design the database to prescribe the knowledge that the LLMs can retrieve and use [\cite{siddharth2024retrieval}]. Finally, an LLM-based agent [\cite{mei2024aios}], [\cite{hong2024data}] is developed to follow the instruction from humans and manipulate structural drawing software like AutoCAD. As validation of the approach, the generation of three different types of structural drawings commonly needed in civil engineering are demonstrated using the proposed method illustrating different approaches for backend logic.

This paper presents a methodology for machine-assisted structural drawing generation utilizing AI. Section 2 includes: (1) an introduction to LLMs, prompt engineering, ReAct, and RAG; (2) the development of the LLM-chain pipeline; (3) the construction of prompts for the LLMs within the pipeline; and (4) an introduction to external background information can be provided. Section 3 comprises: (1) a presentation of the structural drawings to be generated in the case studies; and (2) the specific background information provided for each type of structural drawing. Section 4 features: (1) the output generated by each LLM for different types of structural drawings; and (2) an analysis of these results to illustrate the advantages of the proposed method along with its reliability.

\section{Technical Approach}
\label{sec: technical_approach}
\subsection{Structural Drawing and Relevant Software}
To produce structural drawings, engineers commonly use software like AutoCAD [\cite{autocad2024}], Revit [\cite{revit2024}], and SketchUp [\cite{sketchup2024}], each offering specialized tools and features for different needs. AutoCAD is noteworthy for its extensive adoption and robust capabilities, making it a foundational tool in structural engineering. Given its widespread use and acceptance across the industry, we have chosen AutoCAD for the case studies used in our study.
For the technical implementation, we consider tools that facilitate interaction with AutoCAD. Python libraries such as \texttt{pyautocad}, \texttt{win32com}, etc., provide interfaces to automate and manipulate AutoCAD tasks. Specifically, \texttt{pyautocad} is instrumental for scripting repetitive tasks and generating complex drawings automatically, enhancing productivity and precision in design workflows. Here we utilize the \texttt{pyautocad} library, directing a LLM to generate Python code that interacts directly with AutoCAD.

\subsection{Background: Large Language Model and Prompt Engineering}
\subsubsection{Generative Artificial Intelligence and Large Lanugage Model}
Generative artificial intelligence, encompassing computational methods for creating new, meaningful content such as text, images, or audio, has its roots in the 1950s with the development of Hidden Markov Models [\cite{knill1997hidden}] and Gaussian Mixture Models [\cite{reynolds2015gaussian}]. However, it was the rise of deep learning that marked a significant leap forward in the performance of generative models [\cite{cao2023comprehensive}]. Rapidly, deep generative models branched out across various domains. In NLP, techniques such as N-gram language modeling [\cite{pauls2011faster}], recurrent neural networks (RNNs [\cite{medsker2001recurrent}]), Long Short-Term Memory (LSTM [\cite{graves2012long}]), and Gated Recurrent Units (GRU [\cite{chung2014empirical}]) gained prominence. In contrast, the CV field saw the emergence of Generative Adversarial Networks (GANs [\cite{goodfellow2020generative}]), Variational Autoencoders (VAEs [\cite{kingma2019introduction}]), and diffusion generative models. The transformative moment came with the introduction of the transformer architecture [\cite{vaswani2017attention}], a mechanism employing “self-attention” that has shown superior performance in both CV and NLP tasks [\cite{kingma2019introduction}]. This innovation led to the development of models based on the transformer architecture in various fields. In CV, Vision Transformer (ViT [\cite{han2022survey}]) and Swin Transformer [\cite{liu2021swin}] were created for visual tasks. Concurrently, in NLP, LLMs such as ChatGPT [\cite{chatgpt2025}] and BERT [\cite{devlin2019bert}] have been developed for applications like language translating and question answering. The intersection of generative models across domains, leveraging the transformer architecture, marks a significant stride in the field, underlining the architecture's versatile capabilities.

LLMs are a subset of transformer models, specifically engineered to address Natural Language Processing (NLP) challenges. One distinguishing characteristic of LLMs is their substantial number of parameters, often exceeding a billion. This vast parameter enables them to discern more intricate patterns in data, enhancing their efficacy across various tasks. This feature makes them different than earlier neural network-based NLP models, which typically had far fewer parameters, often in the low millions or fewer. LLMs are categorized based on their architecture into three types: encoder-decoder models like Google's T5 [\cite{gonzalez-adauto2022distraction}], encoder-only models such as BERT [\cite{devlin2019bert}], and decoder-only models, exemplified by ChatGPT [\cite{chatgpt2025}] series. Furthermore, the availability of LLMs ranges from commercial offerings like Claude by Anthropic, to open-source models like Llama series [\cite{touvron2023llama}] developed by Meta. Additionally, LLMs often come in various versions with differing numbers of parameters to suit different tasks and user needs. For instance, Llama is available in versions like 7B, 13B, 33B, and 65B, allowing users to choose a model based on their specific requirements and the complexity of the task at hand. Recently, companies have been exploring wider use of large language models (LLMs) as agents to control various operations of computers and machines. This interest stems from LLMs' ability to generate instructions based on human input. As demonstrated by OpenAI and Anthropic, after receiving instructions from humans, LLMs can assist in searching for information, creating plans, and executing tasks.

\subsubsection{Prompt Engineering}
LLMs may be implemented in three different ways to solve problems: pretraining, fine-tuning and prompt engineering. Pretraining is the initial stage where the model learns general language patterns and knowledge. In this stage, the model is exposed to a wide range of text and the size of the dataset for pretraining is usually in the order of terabytes. Pretraining is typically quite difficult for common users due to this size and the significant computational power that is required. For example, pretraining GPT-3 requires several thousand of Microsoft Azure’s AI-optimized NDv2-series virtual machines (VMs) and each of these VMs contains 8 NVIDIA Tesla V100 GPUs. While the training time for GPT-3 has not been publicly disclosed by OpenAI at this time, it is estimated that it will take several weeks to a few months based on the scale of the model and comparisons with similar models. Thus, for normal users, pretraining is quite expensive and just not feasible. Instead, fine-tuning an LLM refers to the process of starting with a pre-trained model and further training it on a specific, usually smaller dataset to specialize in a particular task or to improve its performance in certain capabilities. This process is critical when one intends to adapt general-purpose language models like GPT-3 or BERT to work well for specific applications. Fine-tuning a pretrained model can reduce the data needed, computational resources required, and training time. However, successfully fine-tuning a pretrained model depends heavily on the fine-tuning dataset, specifically in terms of the data quality and how representative it is for the task at hand. To achieve our goal, the fine-tuning method would require establishing a large dataset that includes a description of each structural drawing paired with the corresponding python code to generate that drawing in AutoCAD. Currently, we lack such information, and fine-tuning is not deemed appropriate.

Drawing from the analysis above, prompt engineering emerges as an appropriate solution for the challenge posed in this work. As an incipient field and technique, prompt engineering involves crafting and refining prompts to derive specific responses or outputs from LLMs [\cite{giray2023prompt}]. These prompts provide LLMs with the necessary information and instructions to adhere to rules, automate processes, and guarantee the desired quality (and quantity) of output. Indeed, prompts serve as a unique form of programming, allowing for tailored outputs and interactions with an LLM [\cite{white2023prompt}]. In fact, various prompt engineering strategies have been developed, including zero-shot prompting [\cite{white2023prompt}], which is viewed as the most straightforward method. Zero-shot prompting requires only the request and essential user input without additional examples or instructions, as shown in Table 1. However, this method may not always fulfill the task satisfactorily due to limitations in task analysis, domain expertise, and example provision. One solution to this limitation is to include examples within the prompt, leading to few-shot prompting, which demonstrates improved performance when models reach a certain size [\cite{touvron2023llama}]. Few-shot prompting, illustrated in Table 1, combines several examples and answers to guide the LLM in performing a specified task, such as sentiment analysis of a sentence. For more tasks requiring more complex reasoning, Chain-of-thought (CoT) prompting [\cite{wei2022chain}] breaks down tasks into smaller, manageable steps, significantly enhancing LLM performance in tasks that require arithmetic, common sense, or symbolic reasoning. Additionally, the ReAct (Reasoning and Acting) framework [\cite{yao2023react}] by using thought-action-observation chain, allows LLMs to generate reasoning traces and task-specific actions, interfacing with external sources for more accurate and factual responses. Other advanced prompting techniques, like Tree-of-Thoughts [\cite{yao2023tree}] and self-consistency [\cite{wang2022self}], further bolster LLM capabilities across diverse tasks, marking significant progress in the field of prompt engineering. 

\begin{table}[htbp]
\centering
\caption{Typical Prompt Engineering Methods}
\renewcommand{\arraystretch}{1.3}
\begin{tabular}{|>{\raggedright\arraybackslash}p{4.5cm}|>{\raggedright\arraybackslash}p{9.5cm}|}
\hline
\textbf{Type of Prompt Engineering} & \textbf{Examples or Sources} \\
\hline
Zero-shot prompting & 
\texttt{Please classify the following text into neutral, negative, or positive based on their sentiment.} \newline
\texttt{Text: I like Hawaii.} \newline
\texttt{Sentiment:} \\
\hline
Few-shot prompting & 
\texttt{This is bad news // Negative} \newline
\texttt{This movie is brilliant // Positive} \newline
\texttt{What a horrible show! // Negative} \newline
\texttt{I like Hawaii! //} \\
\hline
Chain-of-Thought prompting & Wei et al.\ (2022) \cite{wei2022chain} \\
\hline
ReAct prompting & Yao et al.\ (2022) \cite{yao2023react} \\
\hline
Tree of Thoughts & Yao et al.\ (2023) \cite{yao2023tree} \\
\hline
Self-consistency & Wang et al.\ (2022) \cite{wang2022self} \\
\hline
\end{tabular}
\label{tab:tab1}
\end{table}

\subsubsection{Retrieval Augmented Generation}
Pretrained LLMs have demonstrated their capability to encapsulate factual knowledge within their parameters, leading to unparalleled performance in a variety of downstream NLP tasks. Yet, their proficiency in accurately accessing and manipulating this stored knowledge remains constrained [\cite{lewis2020retrieval}]. One significant limitation is their tendency to produce outputs that, while plausible, may not be factually accurate—a phenomenon often referred to as "hallucination" [\cite{lewis2020retrieval}]. Given these challenges, a logical progression is to explore the development of language model-based systems that can tap into external knowledge sources for task completion. This approach not only promises to enhance the factual accuracy of responses but also bolsters the reliability of the outputs, addressing the issue of hallucination by ensuring that the information processed and generated is anchored in verified facts. Meta AI researchers introduced a method called Retrieval Augmented Generation (RAG) to address knowledge-intensive tasks [\cite{lewis2020retrieval}]. RAG takes a user’s input and then retrieves a set of relevant/supporting documents given a source. The LLM does not have direct access to all of the documents rather uses the input as an anchor to retrieve relevant content. Those retrieved documents are concatenated as context with the original input and then fed to the LLMs, which is very useful as LLMs’s parametric knowledge (knowledge stored during pretrain and finetuning) is fixed. RAG allows language models to bypass retraining, enabling access to the latest information for generating reliable outputs. This technique enables the interaction between parametric and non-parametric memories of LLMs [\cite{lewis2020retrieval}], [\cite{gao2023retrieval}]. A workflow showing basic concept of RAG is shown in Figure 1. Based on this concept, different advanced RAG methods are proposed and applied, for example, the pre-retrieval process [\cite{ma2023knowing}] and post-retrieval process [\cite{gao2023retrieval}].

\begin{figure}[htbp]
  \centering
  \includegraphics[width=0.8\textwidth]{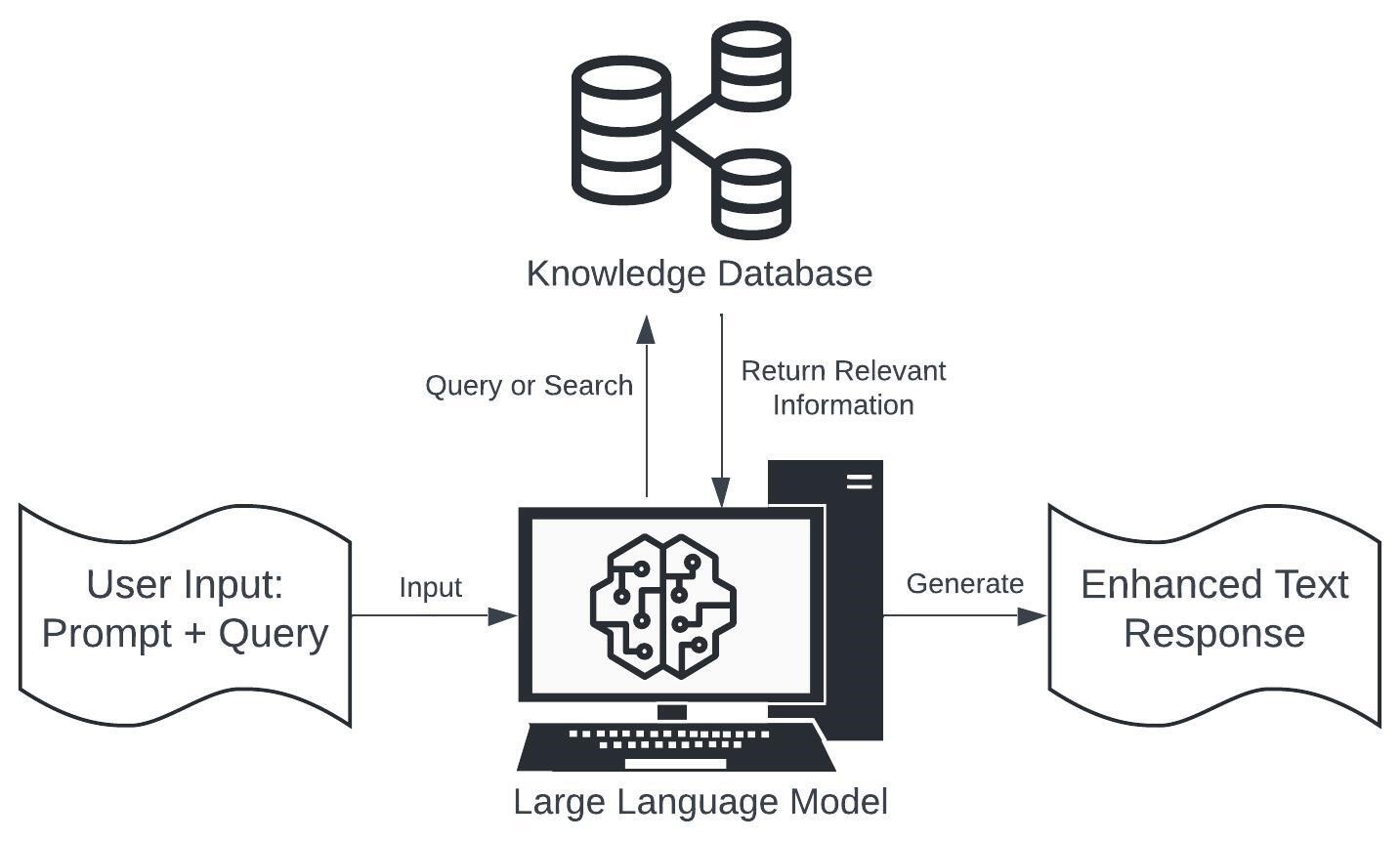}
  \caption{A General Process of RAG}
  \label{fig:fig1}
\end{figure}

\subsection{Design of an LLM-based Structural Drawing Generation Workflow}
\subsubsection{Tasks to be Solved and Challenges}
To establish an LLM-based workflow for generating structural drawings, it's important to delineate the specific tasks that transition natural language descriptions into structured drawings. These tasks encompass the following specific steps:

\begin{itemize}
  \item Identifying the Type of Structural Drawing: This initial step involves determining the specific kind of structural drawing (e.g., concrete beam, steel beam, etc.) needed. Recognizing the drawing type allows for the subsequent use of pertinent information specific to that type of structure.
  \item Refining User's Natural Language Description: Individuals may describe the desired structural drawing in various ways based on their personal knowledge and descriptive habits. It's essential to process and refine these descriptions to extract actionable information.
  \item Acquiring Missing Information: Often, certain details necessary for generating a structural drawing are not explicitly mentioned in the user's description. This might include default parameters or data that require additional calculations. Identifying and procuring this missing information is a key task.
  \item Gathering Additional Information: Some details might pertain directly to the AutoCAD workspace, such as the unit of measurement or whether the file should be saved. Locating this information is also vital.
\end{itemize}

In the LLM-based workflow, these tasks are intended to be performed by LLMs. The result would then be Python code able to interact directly with AutoCAD. This ability requires synthesizing all necessary information, whether computed, processed, or retrieved. Furthermore, once all required data is in place, LLMs should organize this information to enhance the quality of the Python code generated, thereby facilitating a transition from that description to a detailed structural drawing.

Despite significant advancements in the capabilities of LLMs, they still encounter certain limitations that can influence their effectiveness in executing the tasks mentioned previously. These limitations manifest in several ways:

\begin{itemize}
  \item Complex Task Processing: Although LLMs have seen substantial improvements, there are still complex tasks that they find difficult or are unable to complete effectively. This limitation is particularly true for tasks involving multi-step logic or areas outside their “familiar” domains, which can lead to compromised output accuracy and stability.
  \item Task Diversity Handling: The workflow described involves a series of distinct tasks, each requiring different knowledge bases and objectives. Employing a single LLM for all of these tasks may lead to decreased accuracy in the generation and increase the complexity of prompt construction.
  \item Token Size Limitation: While there has been a considerable expansion in the token size limit—for example, the output in models like GPT-4 Turbo now extends to 4096 tokens—this still poses a constraint for extremely complex tasks that require accessing external sources. The token limit can still impact the precision and reliability of the generated content for such intricate tasks.
\end{itemize}

These challenges and limitations highlight the need for the specific design of the LLMs-based workflow to ensure more reliable and accurate outcomes. Thus, in this work, a chain of LLMs method is proposed. In general, there are six steps in this workflow and each step has a LLM to conduct specific work. The detailed setting of those steps is introduced in the following sections.

\subsubsection{Introduction of Step 1: Drawing-type Identification}
\begin{figure}[htbp]
  \centering
  \includegraphics[width=0.8\textwidth]{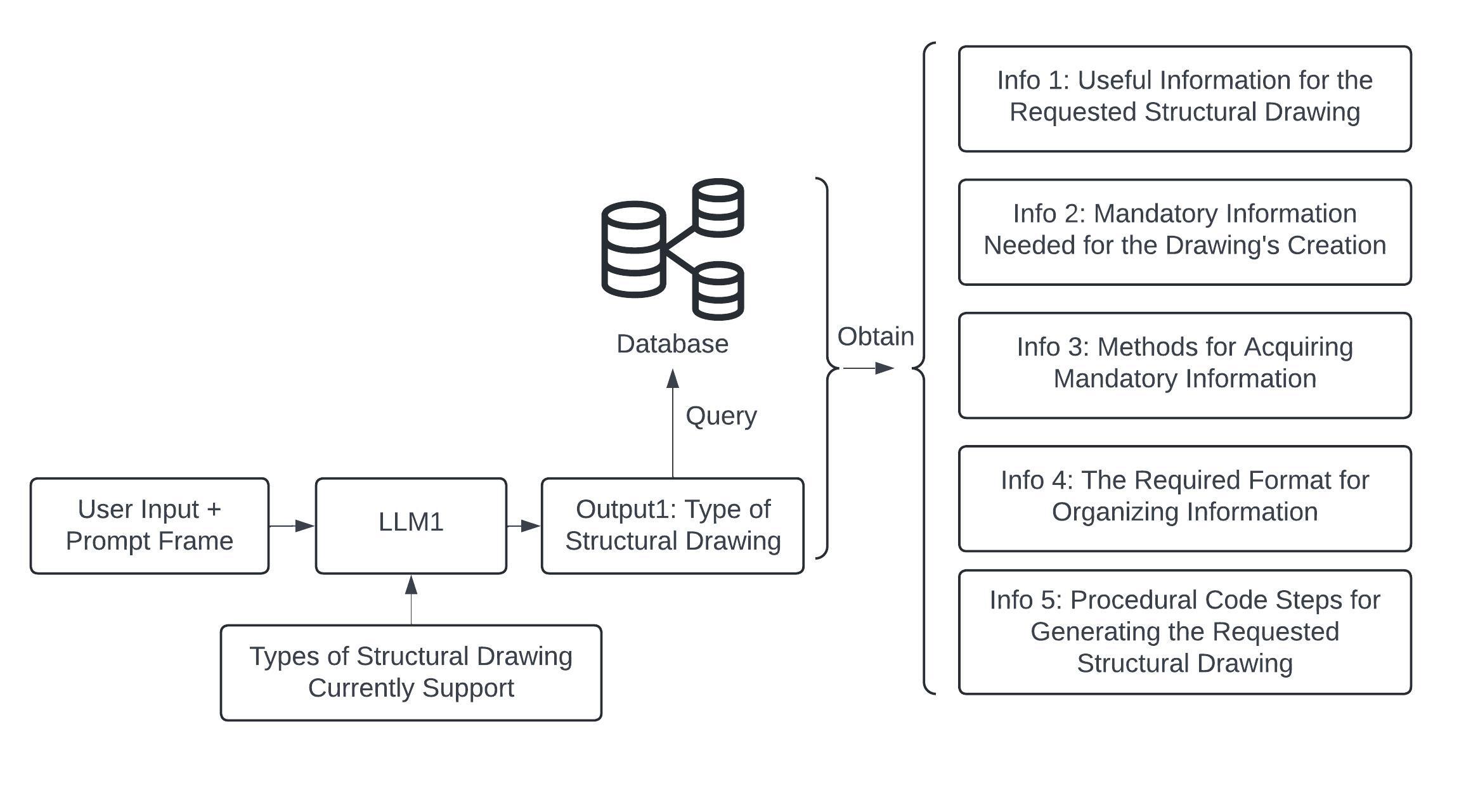}
  \caption{Set up of Step 1}
  \label{fig:fig2}
\end{figure}

As shown in Figure 2, in the initial step a LLM will be deployed to process the user's input (embedded in the specifically designed prompt frame) and discern the specific type of drawings required. To ensure the LLM (designated as LLM1) accurately identifies the drawing type, a critical step to enable subsequent background information retrieval, the prompt will include a list of supported structural drawing types. Once LLM1 successfully determines the drawing type (referred to as output1), we leverage a database curated by human experts to fetch pertinent background details related to the identified structural drawing. This background information encompasses various elements: useful information for the requested structural drawing (Info1), mandatory information needed for the drawing's creation (Info2), methods for acquiring this mandatory information (Info3), the required format for organizing mandatory information (Info4), and the procedural coding steps for generating the requested structural drawing (Info5). All of this information is specifically related to the type of structural drawing requested by the user, indicating that different information will be obtained when a user requests a different type of structural drawing. These pieces of information will play a vital role in the subsequent stages of the workflow, with further specifics on these components provided in later sections. 

The template of the prompt in step 1 is provided in Appendix 1 of this paper. In the prompt, we list the possible types of structural drawings that can be generated and provide general guidelines on how an LLM can differentiate among these types. Additionally, the output format is standardized through the developed prompt. In the thought-action-observation chain, the only work that LLM1 needs to do is find the type of the requested structural drawing. Because the task performed in this step is simple, ChatGPT-3.5 is used for this step.

\subsubsection{Introduction of Step 2: Drawing Details Identification}
\begin{figure}[htbp]
  \centering
  \includegraphics[width=0.8\textwidth]{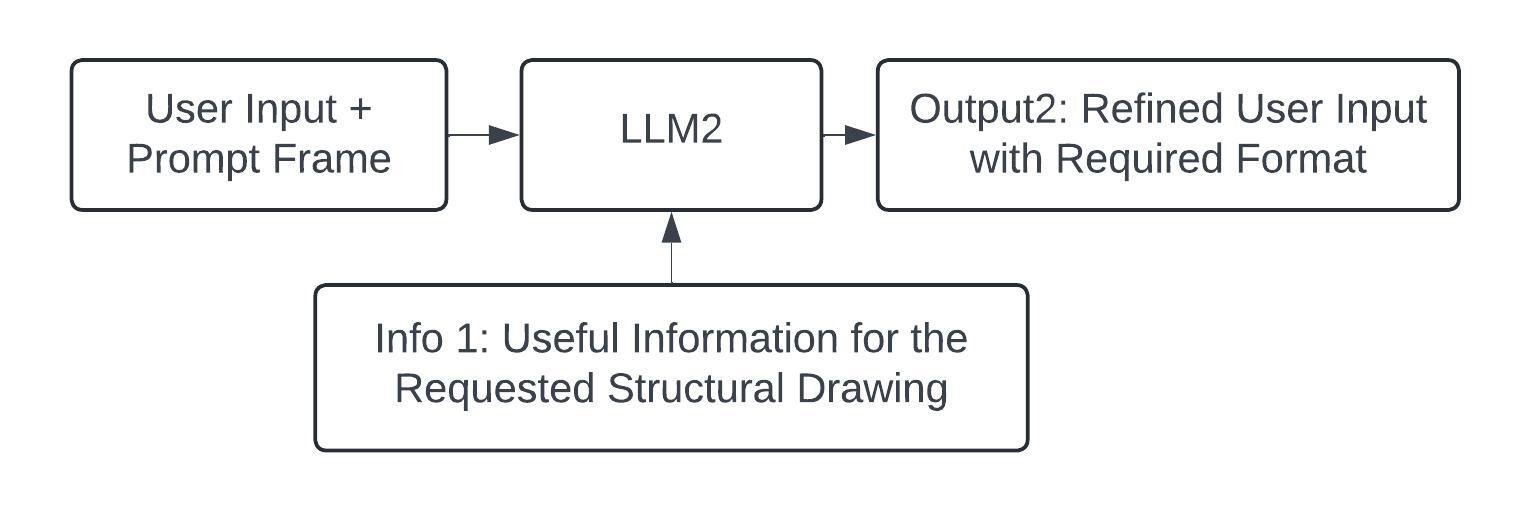}
  \caption{Set up of Step 2}
  \label{fig:fig3}
\end{figure}

There will be considerable variability in how individuals describe the same object to be drawn using natural language. Such differences can lead to inconsistencies in the outputs from an LLM. Therefore, step 2 is designed to refine the user's input and the requirement for refining the input is embedded in the specifically designed prompt). This refinement process involves the LLM determining which parts of the user's input are relevant for generating the structural drawing and which are extraneous. This discernment relies on information (Info1) gathered in step 1. Moreover, the output from step 2 is formatted in a required format, which are stipulated in the prompt. The setting of step 2 and how it processes input is illustrated in Figure 3.

In the prompt for step 2, we incorporate a placeholder, denoted as \texttt{\{useful\_info\}}, which is later replaced with the specific information deemed to be useful for the requested structural drawing. Within the thought-action-observation chain, LLM2 first determines if it recognizes the type of structural drawing requested by the user. Subsequently, LLM2 will parse the user’s input using the designated useful information, which is pre-defined in the database and then used to replace \texttt{\{useful\_info\}} in the prompt. Finally, LLM2 organizes related information under the same bullet point. Details of the prompt used for LLM2 are provided in Appendix 2. Since the task conducted in this step is straightforward, ChatGPT-3.5 is used in this step.

\subsubsection{Introduction of Step 3: Secondary Details}
\begin{figure}[htbp]
  \centering
  \includegraphics[width=0.8\textwidth]{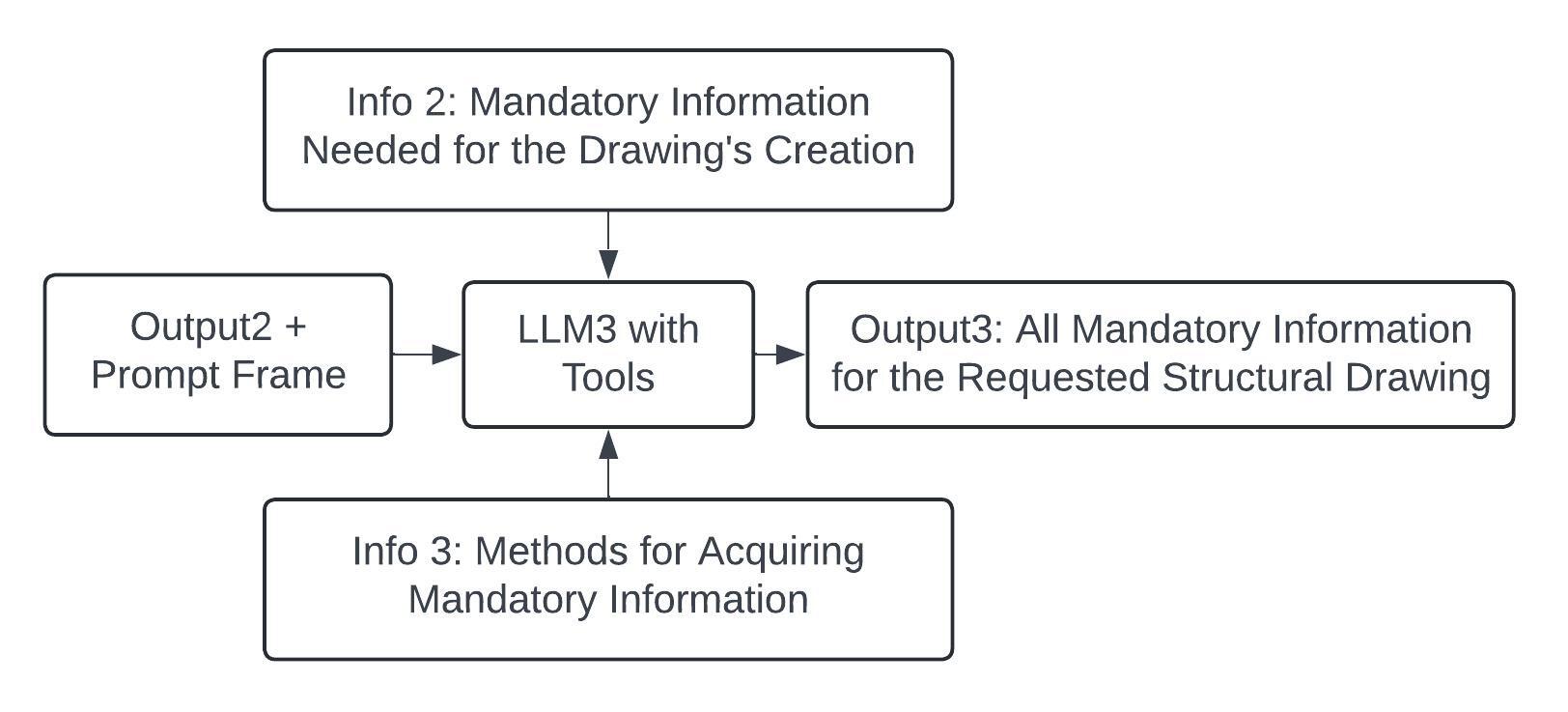}
  \caption{Set up of Step 3}
  \label{fig:fig4}
\end{figure}

After refining the user's input, it is necessary to use another LLM, designated as LLM3, to extract missing information from the refined user input (which is output2 and output2 is embedded in the specifically designed prompt frame for step 3). This necessity arises because some information, such as the position of structural elements, is either defaulted or derivable from other data. Therefore, LLM3 is tasked with acquiring such information in step 3. For this step, two kinds of external sources are utilized. The first kind of external sources include the mandatory information needed to generate the requested structural drawing. The absence of this information would prevent the successful creation of the drawing, making its extraction a primary objective for LLM3. The second kind of external source involves methods for obtaining this mandatory information, such as the calculation procedures needed. These methods equip LLM3 with the necessary understanding to leverage known information to discover what is unknown. Since this step requires mathematical calculations, tools that facilitate addition, subtraction, multiplication, division, and square root operations have been introduced. This enables LLM3 to utilize these tools for more precise computations. By the end of step 3, all mandatory information must be acquired. The setup for this step is depicted in Figure 3. Since the task conducted in this step is complex, ChatGPT-4 is used in this step.

In the prompt for step 3, two placeholders are introduced, marked as \texttt{\{Mandatory\_Info\}} and \texttt{\{background\_sd\}}, which are set to specify the mandatory information required for the drawing’s creation and the methods for acquiring such information, respectively. In the thought-action-observation chain, LLM3 initially identifies reference objects for the mandatory information based on its analysis. Following this, LLM3 explores the relationship, such as geometric relationship, between the mandatory information and these reference objects. In subsequent thought-action-observation sections, LLM3 employs provided calculation tools (multiply, divide, add, subtract, square root) to compute the mandatory information, ensuring all calculations are performed correctly. This step addresses the issue where calculations might have previously been left in a "processing" or "to be completed" status possibly due to a lack of information or laziness of the LLM. Next, LLM3 verifies the accuracy of the arithmetic processes to confirm that all calculations are correct. In the final part of the thought-action-observation chain, LLM3 consolidates all information, including inputs from the user and the data derived in this step. Detailed information on the prompt for this step is provided in Appendix 3.

\subsubsection{Introduction of Step 4: Workspace Details}
\begin{figure}[htbp]
  \centering
  \includegraphics[width=0.8\textwidth]{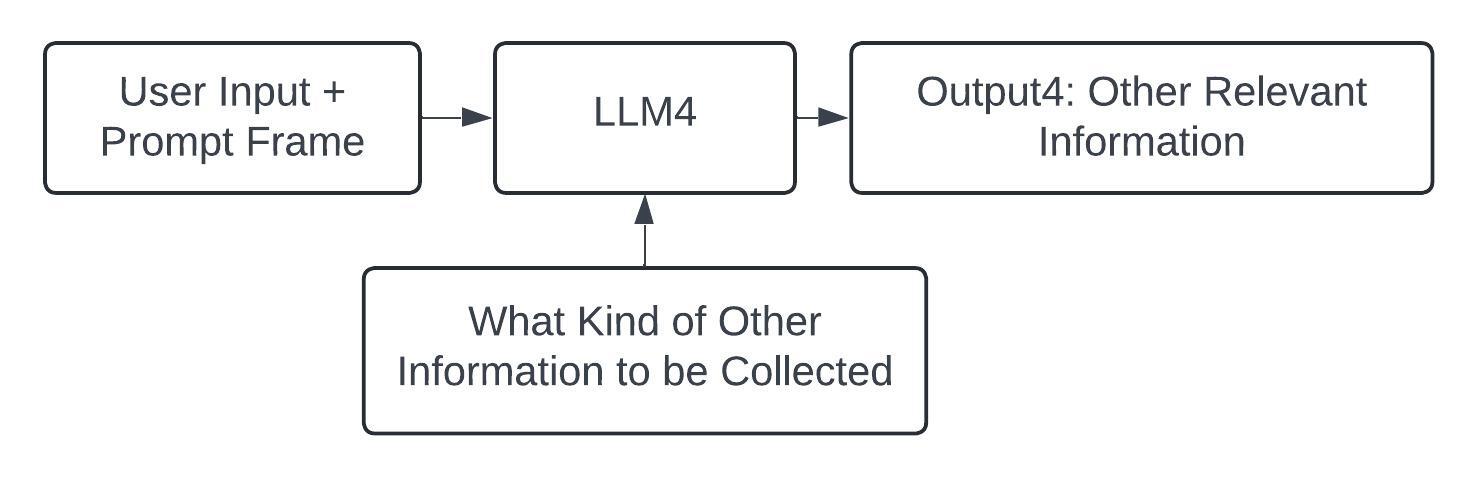}
  \caption{Set up of Step 4}
  \label{fig:fig5}
\end{figure}

To generate structural drawing, additional information that isn't directly related to the objects being drawn may also play a role. For instance, details such as the unit of measurement to be included in the AutoCAD workspace or whether the user intends to save the file after its creation are necessary to ensure that the user obtains the desired output. Step 4 is dedicated to extracting such information using an LLM (named LLM4) from the user’s input (embedded in the specifically designed prompt frame). The configuration and process involved in step 4 are detailed in Figure 5. 

In the prompt for step 4, we instruct LLM4 on the specific additional relevant information that needs to be collected. If this information is not provided by the user, default values should be selected for these relevant parameters. Given the straightforward nature of this task, no detailed thought-action-observation chain has been provided for LLM4. The detailed prompt template is shown in Appendix 4. Since the task conducted in this step is simple, ChatGPT-3.5 is used in this step.

\subsubsection{Introduction of Step 5: Formatting}
\begin{figure}[htbp]
  \centering
  \includegraphics[width=0.8\textwidth]{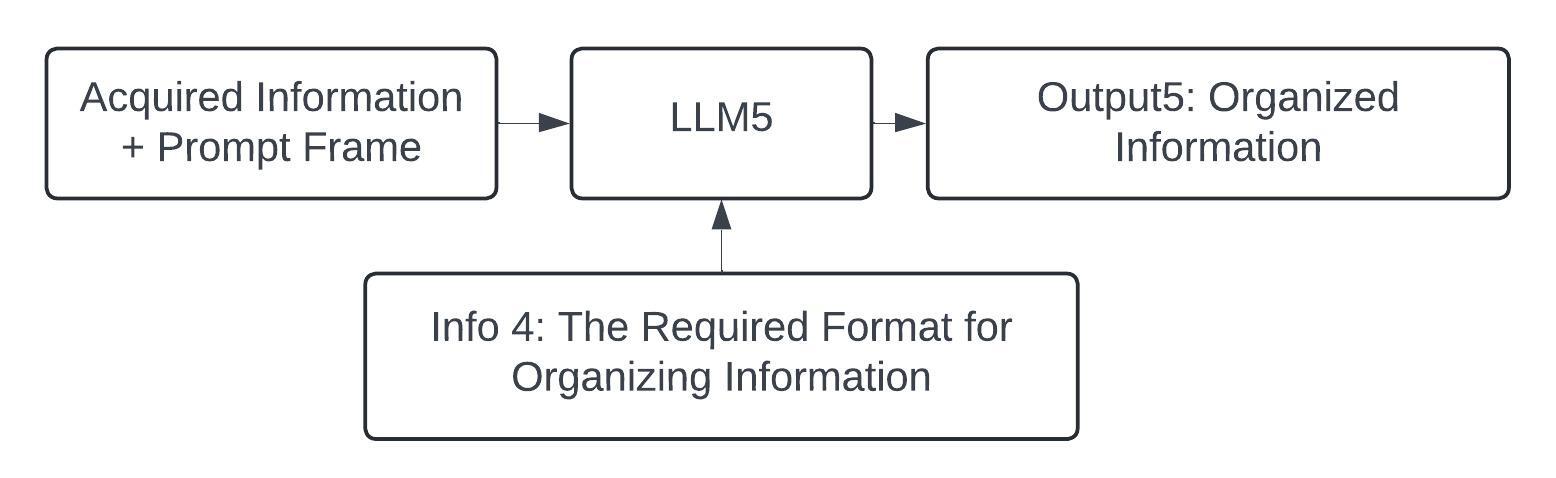}
  \caption{Set up of Step 5}
  \label{fig:fig6}
\end{figure}
After securing all mandatory and relevant information necessary for the requested structural drawing, it is beneficial to arrange this information in a specified format. Providing this organizational structure for the information aids in delineating and documenting the data, facilitating subsequent code generation, which in turn enhances accuracy, reduces instability in the LLM, improve explainability and documentation. Therefore, step 5 involves using an LLM, designated LLM5, to systematically organize all information gathered so far to a specifically designed schema using a JSON format file. The specific format of the JSON that LLM5 is to follow is provided as external knowledge input to ensure effective organization. The details of step 5 are illustrated in Figure 6.

In step 5, the prompt includes a placeholder labeled \texttt{\{JSON\_Requirement\}}, which will be replaced with the specific formatting requirements that all information needs to adhere to. In the first section of the thought-action-observation chain, LLM5 is tasked with recognizing the “keys” and “values” that should be included in the JSON file representing the final format of the requested structural drawing. Following this, in the second section, all the information is transferred and structured into a JSON file, ensuring it meets the specified formatting criteria. The detailed information of the prompt in this step is provided in Appendix 5. Since the task conducted in this step is complex, ChatGPT-4 is used in this step.

\subsubsection{Introduction of Step 6: Code Generation}
\begin{figure}[htbp]
  \centering
  \includegraphics[width=0.8\textwidth]{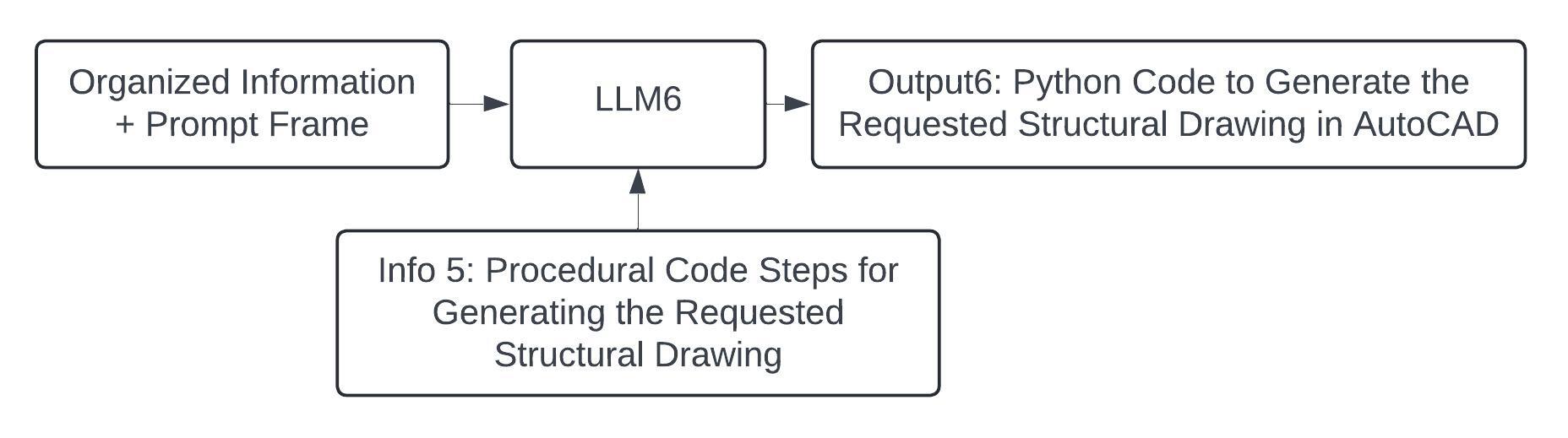}
  \caption{Set up of Step 6}
  \label{fig:fig7}
\end{figure}
The final step involves generating Python code that can interact with the AutoCAD workspace. In this step, an LLM designated as LLM6 utilizes the organized information (output5) as input to create code based on these details. Additionally, this step incorporates external knowledge in the form of procedural code instructions needed for generating the desired structural drawing. These instructions aid LLM6 in understanding the process of crafting code suitable for use in AutoCAD software. The specifics of step 6 are detailed in Figure 7.

In step 6, the thought-action-observation chain is employed to convert the information from the JSON file into Python code. To facilitate this process, instructions on how to generate code using specific Python libraries are provided. In this instance, the \texttt{pyautocad} library is used to interface with AutoCAD. Detailed instructions, such as setting the unit used in the AutoCAD workspace through the Python command \texttt{acad.doc.SetVariable('INSUNITS', unit\_code)} after establishing a connection with \texttt{acad = pyautocad.Autocad()}, are explicitly given to LLM6. This step ensures that LLM6 can accurately follow these directives to generate the necessary code. Within the thought-action-observation chain of step 6, a placeholder labeled \texttt{\{steps\}} is included, which will later be replaced with the precise steps needed to generate the specified structural drawing. The details are shown in Appendix 6. Since the task conducted in this step is complex, ChatGPT-4 is used in this step.

Based on the developed approach described in the previous sections, the comprehensive workflow adopted for this research is depicted in Figure 8. This visualization outlines the sequence and integration of the various steps involved in the process. This workflow is modular, assigning specific tasks to individual LLMs, thereby allowing users to easily reuse, replace, modify, and debug these modules for different objectives. In this work, the proposed workflow is constructed using libraries provided by LangChain, which facilitates the development and integration of language model capabilities into the overall system [\cite{das2024llm}]. The specific setting for different types of structural drawing will be introduced in section 3. 
\begin{figure}[htbp]
  \centering
  \includegraphics[width=0.8\textwidth]{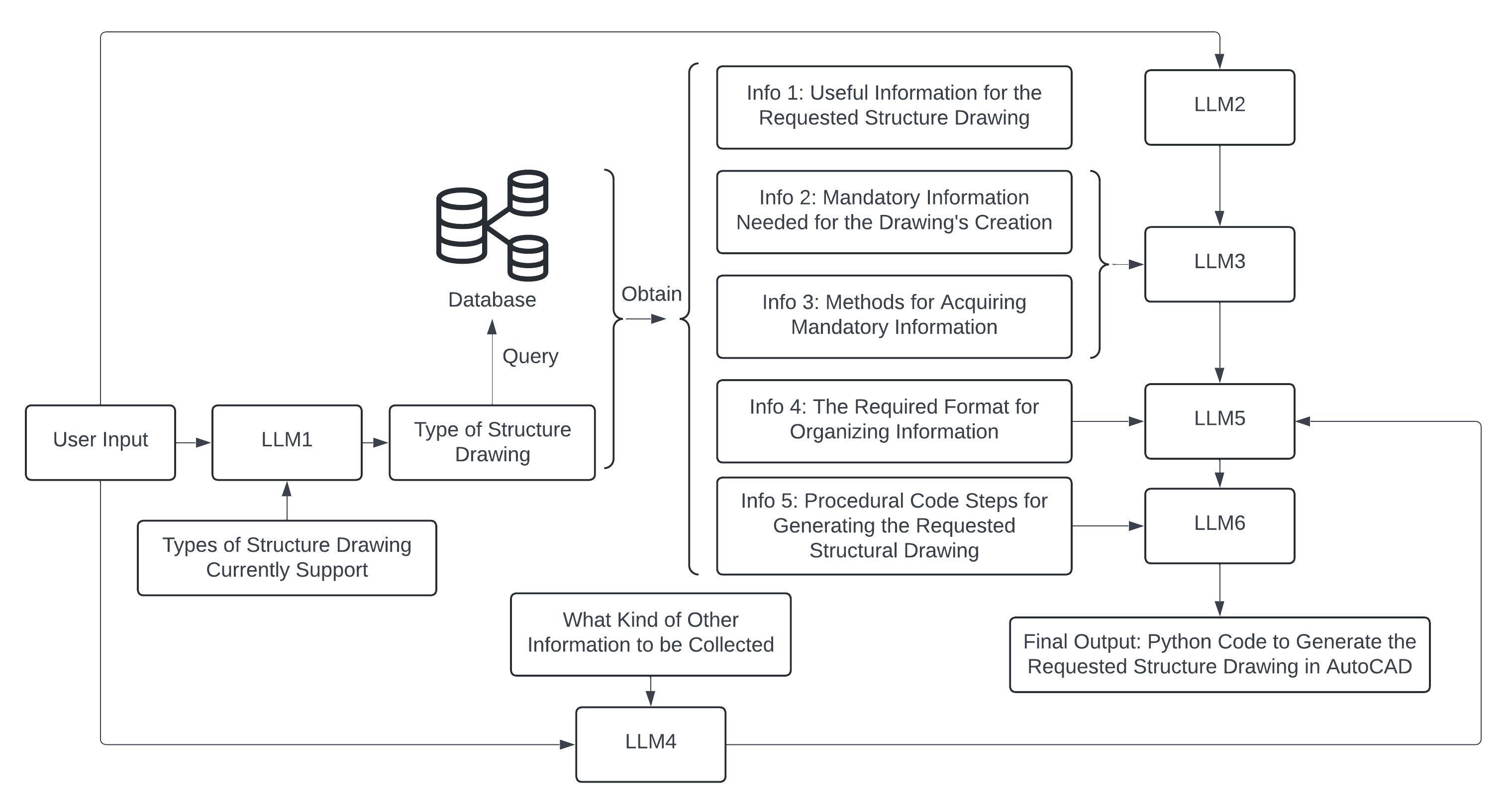}
  \caption{Overall Workflow for LLM Agent}
  \label{fig:fig8}
\end{figure}

\section{Illustrative Case Studies}
\label{sec:illustrative_case_studies}
\subsection{Drawing of Rectangular Reinforced Concrete Beam Cross-section}
\subsubsection{Logic of the Case Study}
In general, the structural design of reinforced concrete members is focused on determining the required steel reinforcement for a selected cross-section based on the loading conditions, the structure type, geometric constraints, minimum design requirements, and past experience. Despite this design process being inherently technical, multiple viable solutions exist, requiring the engineer’s judgment to select the most suitable solution. The final design often requires multiple iterations to resolve conflicts between members, optimize material usage – such as rebar usage – and gradually incorporate design specifications. In this context, LLMs serve as a tool for the structural engineer to quickly create structural drawings or update them multiple times, streamlining the design process. 
\subsubsection{Specific Workflow for this Case Study}
As explained in section 2.3, in the first step of the workflow LLM1 identifies the type of section to process which in this case is reinforced concrete. Next, LLM2 identifies and collects the relevant information from the user’s prompts. In the case of basic reinforced concrete design, the information to be considered as input includes the cross-section size, bar sizes, number of bars per layer, spacing between layers, and clear cover width. 

LLM3 determines what information from this list is mandatory to generate the section with a Python code. The information can be provided in the prompt in two ways: explicitly, where the user specifies the coordinates directly, or, more commonly, descriptively, where the coordinates will be calculated by a list of equations that the framework is equipped to compute. In this case study, mandatory information corresponds to (1) the height and depth of the cross-section to get geometric features such as vertices and side lengths, (2) the clear cover thickness, (3) stirrups details such as rebar size and transverse spacing to draw the bending radius of the stirrups for the vertices and hooks, (4) the longitudinal rebar size, (5) number of bars per layer and (6) number of layers. An example of not explicitly stated mandatory information from received data is the diameter of a No 8 rebar. At the end of the process, the geometrical features for each of the design components will be defined through the coordinates of vertices in the format “(x, y)", where x and y are coordinates in the AutoCAD workspace. Regarding the construction of the drawing, due to the output limitations of the LLM (GPT-4), step 3 is divided into three sub-steps: acquiring mandatory information except for stirrup information, acquiring information of internal lines, external lines, and arc lines, and acquiring information of lines for stirrup hooks. However, the template for the prompt for these sub-steps is the same, and in the future, if the power of the LLM increases, a single step can complete this process.

LLM4 is focused on identifying what other relevant information needs to be collected to complete the task of preparing code for the drawings. In this case study, LLM4 also (i) determines or defaults the unit of the structural drawing, and (ii) understands whether or not the user wishes to save the file.

LLM5 then organizes all the information of the requested structural drawing into the required format, simplifying the analytical complexity of the final step—generating Python code. Thus, all information is stored in a JSON file format. The external knowledge outlines the “keys” and “values” in the JSON file required for the reinforced concrete beam cross-section, which includes whether the drawing file needs to be saved, the unit of measurement used, the specific drawing features such as the Coordinates of the cross-section vertices, rebar centers, and radii/diameters, endpoints of internal and external stirrup lines, arc segments, and hook lines for the stirrups.

Finally, LLM6 is responsible for generating Python code based on the information generated and organized from LLM5. The steps for generating this code are provided as external knowledge for LLM6. The first step involves setting units with the Python command \texttt{acad.doc.SetVariable('INSUNITS', unit\_code)}. Then, vertices of the reinforced concrete beam cross-section are created with \texttt{"pyautocad.APoint(vertices\_coordinate)"}. Sides are drawn using \texttt{"acad.model.AddLine(pyautocad.APoint(end1), pyautocad.APoint(end2))"}. Steel rebar is generated through \texttt{"acad.model.AddCircle(pyautocad.APoint(center), radius)"}. The internal lines, arcs, and hook lines are created with \texttt{"acad.model.AddLine(pyautocad.APoint(end1), pyautocad.APoint(end2))"}, \texttt{"acad.model.AddArc(pyautocad.APoint(center\_x, center\_y), radius, start\_angle, end\_angle))"}, and \texttt{" acad.model.AddLine(pyautocad.APoint(end1),pyautocad.APoint(end2))"}, respectively. More details on the code are in the Appendix 7-11. 

\subsection{Drawing for Steel Beam Cross-section}
\subsubsection{Logic of the Case Study}
Unlike reinforced concrete design, the dimensions of most steel sections are predefined from the supplier’s catalog, typically standardized to comply with the specifications set forth by the American Institute for Steel Construction (AISC) [\cite{aiscmanual15}]. In this case study, the LLM will analyze the user’s input and retrieve the predefined steel beam cross-section from a file system or database. In other cases, such as steel built-up sections or specifically designed steel beams, the method in the concrete case study can be used.   

\subsubsection{Specific Workflow for this Case Study}
In this case study, the useful information for LLM2 includes the specific type of steel beam cross-section and the position in the AutoCAD workspace where the structural drawing should be placed. The position is optional in this framework and can be used to connect the generated CAD drawing to specific locations in a structural drawing sheet. For example, in the prompt the user provides HP360x174, dictating the type of steel cross-section to be retrieved, along with the coordinates to position the section. Consequently, the mandatory information needed for LLM3 encompasses the section information and the coordinates of the bottom left vertex of the steel beam cross-section. In the absence of the coordinates, the default coordinates for the bottom left vertex are set to (0, 0). Alternatively, the user can include the position of other parts of the cross-section and ask the LLM to determine the coordinate of the bottom left vertex from geometric information. 

Next, LLM4 determines the units and assesses whether the user intends to save the file post-generation. LLM5 then organizes the information into a JSON file format, which includes keywords such as whether the file should be saved, the units, the type of cross-section for the drawing, the specified steel beam cross-section, and the position of the bottom left vertex of the steel beam cross-section. 

Finally, organized information is provided to LLM6 for generating the python code. This information includes: use the type of the steel beam cross-section to determine the source file path with the Python command \texttt{"os.path.join(path\_to\_folder, type\_of\_steel\_beam\_cross-section + '.dwg')"}, open the source file using \texttt{"acad.app.Documents.Open(source\_file\_path)"}, pause the program for 1 second to ensure the connection has been successfully established using \texttt{"time.sleep(1)"}, set the active document to the source file using \texttt{"acad.app.ActiveDocument = source\_document"}, select all objects in the source file with \texttt{"acad.app.ActiveDocument.SendCommand('SELECT ALL\\n')"}, copy all selected objects with \texttt{"acad.app.ActiveDocument.SendCommand('COPYCLIP\\n')"}, instruct the path for the target file using \texttt{"os.path.join(os.getcwd(), 'targetfile.dwg')"}, open the target file using \texttt{"target\_document = acad.app.Documents.Open(target\_file)"}, set the active document to the target file with \texttt{"acad.app.ActiveDocument = target\_document"}, paste the objects into the target file at the specified position using \texttt{"acad.app.ActiveDocument.SendCommand('PASTECLIP ' + str(x\_coordinate) + ',' + str(y\_coordinate) + '\\n')"}, close the source document with \texttt{"source\_document.Close()"}. The details of the external information used in this case study is provided in Appendix 7-11

\subsection{Drawing for Precast Concrete Beam Corss-section}
\subsubsection{Logic of the Case Study}
Typically, precast concrete sections are predefined in suppliers’ catalogs and design manuals. For example, Chapter 406 of the 2013 design manual from the Indiana Department of Transportation defines the sizes for different types of precast beams. However, the number of strands must be specified by the structural engineer. Thus, the method for creating precast beam cross-sections incorporates techniques from the first two case studies, as detailed in sections 3.1 and 3.2. In this case study, the LLM first retrieves the predetermined precast beam cross-section and then modifies the strand openings based on the user's specifications.

\subsubsection{Specific Workflow for this Case Study}
In this case study, the useful information for LLM2 includes the type of predetermined precast beam cross-section, the position of the specific cross-section, and the number of strands the user intends to use. The mandatory information for LLM3 encompasses the position of the bottom left vertex (default to (0, 0)) and the positions of the strands. Methods for acquiring this mandatory information involve a general hint that the position of the left vertex can be derived from its geometric relationships, along with instructions on determining the precise positions of the strands based on their potential locations and the number requested by the user. 

Next, LLM4 determines the units of the structural drawing and whether the user wishes to save the file after generation. LLM5 compiles all the useful information, mandatory information, and other information from the user’s request and previous steps into a specified JSON file. This file includes “keys” for the position of the left vertex and the centers of each strand.

Finally, LLM6 generates Python code based on the organized information. The steps included in LLM6 are: use category and subcategory of the prestressed beam cross-section to determine the source file path (command: \texttt{os.path.join(os.getcwd(),'Preset\_Prestressed\_Concrete','category\_subcategory.dwg'))}. Remember to replace \texttt{'category'} and \texttt{'subcategory'} in \texttt{'category\_subcategory.dwg'} with requested category and subcategory of the prestressed concrete input by the user; sleep python code for 1 seconds. (command: \texttt{time.sleep(1))}; activate source file. (command: \texttt{acad.app.ActiveDocument = source\_document)}; select all objects in the source file using prompt to AutoCAD software. (command: \texttt{acad.app.ActiveDocument.SendCommand('SELECT ALL  '))}; copy all selected objects. (command: \texttt{acad.app.ActiveDocument.SendCommand('COPYCLIP '))}; build the path for target file (command: \texttt{os.path.join(os.getcwd(), 'targetfile.dwg'))}; build the connection with the target file. (command: \texttt{target\_document = acad.app.Documents.Open(target\_file))}; activate target file. (command: \texttt{acad.app.ActiveDocument = target\_document)}; paste the objects to target file in the requested position. (command: \texttt{acad.app.ActiveDocument.SendCommand('PASTECLIP x\_coordinate,y\_coordinate '))}; close source document. (command: \texttt{source\_document.Close())}; draw strands: find coordinates of the center and radius of each strands. (command: \texttt{acad.model.AddCircle(pyautocad.APoint(center), radius))}. The details of the external information used in this case study are provided in Appendix 7-11.

\section{Case Study Results}
\label{sec:case_study_results}
\subsection{Generation Results for Reinforced Concrete BEam Cross-section}
The generated text varies between trials, yet the essential components, particularly those that impact the final structural drawing, remain consistent. The example outcomes for each step are displayed in Table 2 below, using the example input: “I would like to draw a 24x14in RC cross-section with No 4 closed stirrups at 5 in. It needs three rebar layers. The top and bottom layer will have 4 No 8 and 2 No 4, respectively. The middle layer will have 2 No 4. Consider a 2in clear cover.”. In practice, the input is drafted by the user based on their specific design; the result here is just shown as an example. Please note that Table 2 only includes the final outcomes; a comprehensive breakdown of the thought-action-observation analysis process is available in Appendix 7-10.

\begin{longtable}{|c|p{12cm}|}
\caption{Final results for reinforced concrete beam cross-section}  \label{tab:tab2} \\
\hline
\textbf{No. of Steps} & \textbf{Final Results} \\
\hline
\endfirsthead

\hline
\textbf{No. of Steps} & \textbf{Final Results (continued)} \\
\hline
\endhead

1 & The type of structural drawing is: rectangular concrete beam cross-section \\
\hline
2 & Final Answer: \newline
<result> \newline
- Type of Structure: rectangular concrete beam cross-section \newline
- Height of cross-section: 24in \newline
- Width of cross-section: 14in \newline
- Number of rebars: 3 layers \newline
- Rebar information: \newline
\quad - Top layer: 4 No 8 \newline
\quad - Middle layer: 2 No 4 \newline
\quad - Bottom layer: 2 No 4 \newline
- Stirrup information: No 4 \newline
- Thickness of clear cover: 2in \newline
</result> \\
\hline
3-1 & Final Answer: \newline
<result> \newline
- bottom left vertex: (0, 0) \newline
- top left vertex: (0, 24) \newline
- top right vertex: (14, 24) \newline
- bottom right vertex: (14, 0) \newline
- left appearance: x=0 \newline
- top appearance: y=24 \newline
- right appearance: x=14 \newline
- bottom appearance: y=0 \newline
- Height: 24in \newline
- Width: 14in \newline
- Radius of Stirrup: 0.25in \newline
- Diameter of Stirrup: 0.5in \newline
- Thickness of clear cover: 2in \newline
- Total number of all rebars: 8 \newline
- Coordinates and radius of the center of each rebar: \newline
\quad - Top layer: [(3in, 21in), 0.5in)], [(5.6667in, 21in), 0.5in)], [(8.3333in, 21in), 0.5in)], [(13in, 21in), 0.5in)] \newline
\quad - Middle layer: [(2.75in, 11.875in), 0.25in)], [(11.25in, 11.875in), 0.25in)] \newline
\quad - Bottom layer: [(2.75in, 2.75in), 0.25in)], [(11.25in, 2.75in), 0.25in)] \newline
</result> \\
\hline
3-2 & Final Answer: \newline
<result> \newline
- Radius of Stirrup: 0.25in \newline
- Diameter of Stirrup: 0.5in \newline
- C1: [(3in, 21in), 0.5in] \newline
- C2: [(11in, 21in), 0.5in] \newline
- C3: [(11.25in, 2.75in), 0.25in] \newline
- C4: [(2.75in, 2.75in), 0.25in] \newline
- L1: [(2.5, 21), (2.5, 2.75)] \newline
- L2: [(3.9142, 21.5), (11, 21.5)] \newline
- L3: [(11.5, 21), (11.5, 2.75)] \newline
- L4: [(2.75, 2.5), (11.25, 2.5)] \newline
- L5: [(2, 21), (2, 2.75)] \newline
- L6: [(3, 22), (11, 22)] \newline
- L7: [(12, 21), (12, 2.75)] \newline
- L8: [(2.75, 2), (11.25, 2)] \newline
- A1: [3, 21, 1, 45, 180] \newline
- A2: [11, 21, 1, 0, 90] \newline
- A3: [11.25, 2.75, 0.75, 270, 0] \newline
- A4: [2.75, 2.75, 0.75, 180, 270] \newline
- Length of Hook: 3 inches \newline
</result> \\
\hline
3-3 & Final Answer: \newline
<result> \newline
- Lh1: [(3.7071, 21.7071), (5.8284, 19.5858)] \newline
- Lh2: [(3.3536, 21.3536), (5.4749, 19.2322)] \newline
- Lh3: [(5.8284, 19.5858), (5.4749, 19.2322)] \newline
- Lh4: [(2.6464, 20.6464), (4.7678, 18.5251)] \newline
- Lh5: [(2.5, 20.0858), (4.4142, 18.1716)] \newline
- Lh6: [(4.7678, 18.5251), (4.4142, 18.1716)] \newline
</result> \\
\hline
4 & <result> \newline
Save: False \newline
Unit: Inch \newline
</result> \\
\hline
5 & Final Answer: \newline
<result> \newline
\{ \newline
\quad "Save": false, \newline
\quad "Unit": "Inch", \newline
\quad "Type of Structural Drawing": "rectangular concrete beam cross-section", \newline
\quad "Coordinates of Four Vertices": \{ \newline
\quad\quad "bottom left": [0, 0], \newline
\quad\quad "top left": [0, 24], \newline
\quad\quad "top right": [14, 24], \newline
\quad\quad "bottom right": [14, 0] \newline
\quad \}, \newline
\quad "End Point of Four Sides": \{ \newline
\quad\quad "left": \{ "end1": [0, 0], "end2": [0, 24] \}, \newline
\quad\quad "top": \{ "end1": [0, 24], "end2": [14, 24] \}, \newline
\quad\quad "right": \{ "end1": [14, 24], "end2": [14, 0] \}, \newline
\quad\quad "bottom": \{ "end1": [14, 0], "end2": [0, 0] \} \newline
\quad \}, \newline
\quad "Center of Rebars": [ \newline
\quad\quad [3, 21], [5.6667, 21], [8.3333, 21], [11, 21], \newline
\quad\quad [2.75, 11.875], [11.25, 11.875], [2.75, 2.75], [11.25, 2.75] \newline
\quad ], \newline
\quad "Radius of Rebars": [ \newline
\quad\quad 0.5, 0.5, 0.5, 0.5, \newline
\quad\quad 0.25, 0.25, 0.25, 0.25 \newline
\quad ], \newline
\quad "Radius and Diameter of Stirrup": [0.25, 0.5], \newline
\quad "End Points of Internal and External Lines of Stirrup": \{ \newline
\quad\quad "L1": \{ "end1": [2.5, 21], "end2": [2.5, 2.75] \}, \newline
\quad\quad "L2": \{ "end1": [3.9142, 21.5], "end2": [11, 21.5] \}, \newline
\quad\quad "L3": \{ "end1": [11.5, 21], "end2": [11.5, 2.75] \}, \newline
\quad\quad "L4": \{ "end1": [2.75, 2.5], "end2": [11.25, 2.5] \}, \newline
\quad\quad "L5": \{ "end1": [2, 21], "end2": [2, 2.75] \}, \newline
\quad\quad "L6": \{ "end1": [3, 22], "end2": [11, 22] \}, \newline
\quad\quad "L7": \{ "end1": [12, 21], "end2": [12, 2.75] \}, \newline
\quad\quad "L8": \{ "end1": [2.75, 2], "end2": [11.25, 2] \} \newline
\quad \}, \newline
\quad "Arc Lines of Stirrup": \{ \newline
\quad\quad "A1": [3, 21, 1, 45, 180], \newline
\quad\quad "A2": [11, 21, 1, 0, 90], \newline
\quad\quad "A3": [11.25, 2.75, 0.75, 270, 0], \newline
\quad\quad "A4": [2.75, 2.75, 0.75, 180, 270] \newline
\quad \}, \newline
\quad "Hook lines of Stirrup": \{ \newline
\quad\quad "Lh1": \{ "end1": [3.7071, 21.7071], "end2": [5.8284, 19.5858] \}, \newline
\quad\quad "Lh2": \{ "end1": [3.3536, 21.3536], "end2": [5.4749, 19.2322] \}, \newline
\quad\quad "Lh3": \{ "end1": [5.8284, 19.5858], "end2": [5.4749, 19.2322] \}, \newline
\quad\quad "Lh4": \{ "end1": [2.6464, 20.6464], "end2": [4.7678, 18.5251] \}, \newline
\quad\quad "Lh5": \{ "end1": [2.5, 20.0858], "end2": [4.4142, 18.1716] \}, \newline
\quad\quad "Lh6": \{ "end1": [4.7678, 18.5251], "end2": [4.4142, 18.1716] \} \newline
\quad \} \newline
\} \newline
</result> \\
\hline

6 & \begingroup\raggedright\footnotesize
<result> \newline
from pyautocad import Autocad, APoint \newline
from math import radians \newline

acad = Autocad() \newline
acad.doc.SetVariable('INSUNITS', 1) \newline

vertices = \{ \newline
\quad "bottom\_left": APoint(0, 0), \newline
\quad "top\_left": APoint(0, 24), \newline
\quad "top\_right": APoint(14, 24), \newline
\quad "bottom\_right": APoint(14, 0) \newline
\} \newline

sides = \{ \newline
\quad "left": acad.model.AddLine(APoint(0, 0), APoint(0, 24)), \newline
\quad "top": acad.model.AddLine(APoint(0, 24), APoint(14, 24)), \newline
\quad "right": acad.model.AddLine(APoint(14, 24), APoint(14, 0)), \newline
\quad "bottom": acad.model.AddLine(APoint(14, 0), APoint(0, 0)) \newline
\} \newline

rebars = [ \newline
\quad acad.model.AddCircle(APoint(3, 21), 0.5), \newline
\quad acad.model.AddCircle(APoint(5.6667, 21), 0.5), \newline
\quad acad.model.AddCircle(APoint(8.3333, 21), 0.5), \newline
\quad acad.model.AddCircle(APoint(11, 21), 0.5), \newline
\quad acad.model.AddCircle(APoint(2.75, 11.875), 0.25), \newline
\quad acad.model.AddCircle(APoint(11.25, 11.875), 0.25), \newline
\quad acad.model.AddCircle(APoint(2.75, 2.75), 0.25), \newline
\quad acad.model.AddCircle(APoint(11.25, 2.75), 0.25) \newline
] \newline

stirrup\_lines = \{ \newline
\quad "L1": acad.model.AddLine(APoint(2.5, 21), APoint(2.5, 2.75)), \newline
\quad "L2": acad.model.AddLine(APoint(3.9142, 21.5), APoint(11, 21.5)), \newline
\quad "L3": acad.model.AddLine(APoint(11.5, 21), APoint(11.5, 2.75)), \newline
\quad "L4": acad.model.AddLine(APoint(2.75, 2.5), APoint(11.25, 2.5)), \newline
\quad "L5": acad.model.AddLine(APoint(2, 21), APoint(2, 2.75)), \newline
\quad "L6": acad.model.AddLine(APoint(3, 22), APoint(11, 22)), \newline
\quad "L7": acad.model.AddLine(APoint(12, 21), APoint(12, 2.75)), \newline
\quad "L8": acad.model.AddLine(APoint(2.75, 2), APoint(11.25, 2)) \newline
\} \newline

stirrup\_arcs = \{ \newline
\quad "A1": acad.model.AddArc(APoint(3, 21), 1, radians(45), radians(180)), \newline
\quad "A2": acad.model.AddArc(APoint(11, 21), 1, radians(0), radians(90)), \newline
\quad "A3": acad.model.AddArc(APoint(11.25, 2.75), 0.75, radians(270), radians(0)), \newline
\quad "A4": acad.model.AddArc(APoint(2.75, 2.75), 0.75, radians(180), radians(270)) \newline
\} \newline

stirrup\_hooks = \{ \newline
\quad "Lh1": acad.model.AddLine(APoint(3.7071, 21.7071), APoint(5.8284, 19.5858)), \newline
\quad "Lh2": acad.model.AddLine(APoint(3.3536, 21.3536), APoint(5.4749, 19.2322)), \newline
\quad "Lh3": acad.model.AddLine(APoint(5.8284, 19.5858), APoint(5.4749, 19.2322)), \newline
\quad "Lh4": acad.model.AddLine(APoint(2.6464, 20.6464), APoint(4.7678, 18.5251)), \newline
\quad "Lh5": acad.model.AddLine(APoint(2.5, 20.0858), APoint(4.4142, 18.1716)), \newline
\quad "Lh6": acad.model.AddLine(APoint(4.7678, 18.5251), APoint(4.4142, 18.1716)) \newline
\} \newline
</result>
\par\endgroup \\ 
\hline

\end{longtable}

\begin{figure}[htbp]
  \centering
  \includegraphics[width=0.3\textwidth]{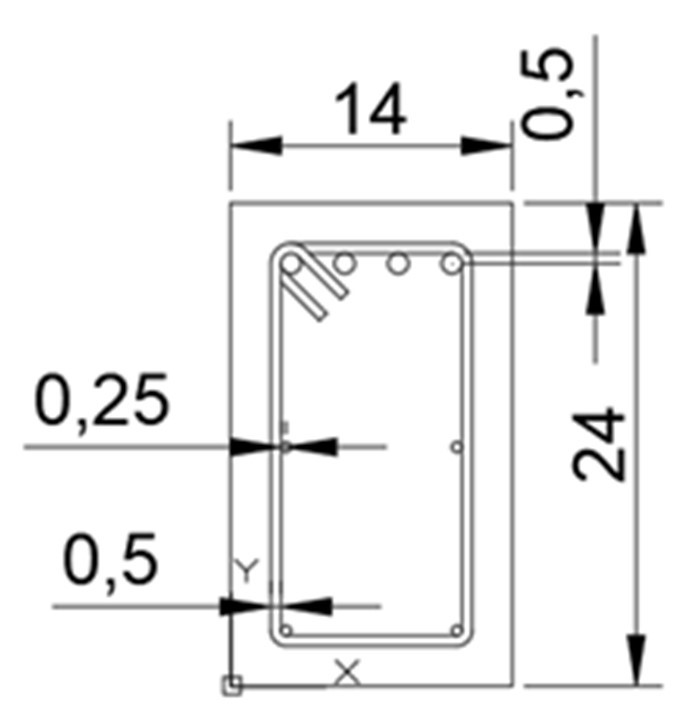}
  \caption{Concrete beam cross section sample (annotations are added by human not LLMs)}
  \label{fig:fig9}
\end{figure}

\subsection{Generation Results for Steel Beam Cross-section}
The example outcomes of each step related to the steel beam cross-section are presented in Table3, based on the input “I would like to draw W1100X390”. Again, this is just an example input, and users can provide similar prompts. The sample drawing is shown in figure 10 and since we directly copy the drawing from source files, users can regulate source files based on their need. The sample drawing is simplified just for demonstration.

\begin{longtable}{|c|p{12cm}|}
\caption{Final results for each step for steel beam cross-section}  \label{tab:tab3} \\
\hline
\textbf{No. of Steps} & \textbf{Final Results} \\
\hline
\endfirsthead

\hline
\textbf{No. of Steps} & \textbf{Final Results (continued)} \\
\hline
\endhead
1 & The type of structural drawing is: steel beam cross-section \\
\hline
2 & Final Answer: \newline
<result> \newline
- Type of Structure: steel beam cross-section \newline
- Steel Beam Cross-section: W1100X390 \newline
</result> \\
\hline
3 & Final Answer: \newline
<result> \newline
Type of Steel Beam Cross-section: W1100X390 \newline
Coordinate of Bottom Left of the Cross-section: (0, 0) \newline
</result> \\
\hline
4 & <result> \newline
Save: False \newline
Unit: Millimeter \newline
</result> \\
\hline
5 & Final Answer: \newline
<result> \newline
\{ \newline
\quad "Save": false, \newline
\quad "Unit": "Millimeter", \newline
\quad "Type of Structural drawing": "steel beam cross-section", \newline
\quad "Type of the requested steel beam cross-section": "W1100X390", \newline
\quad "Position of the bottom left of the steel beam cross-section": "(0, 0)" \newline
\} \newline
</result> \\
\hline
6 & \begingroup\raggedright\footnotesize
<result> \newline
import pyautocad \newline
import os \newline
import time \newline

acad = pyautocad.Autocad() \newline

source\_file\_path = os.path.join(os.getcwd(), 'steelBeamDrawingSet', 'W1100X390.dwg') \newline
source\_document = acad.app.Documents.Open(source\_file\_path) \newline
time.sleep(1) \newline
acad.app.ActiveDocument = source\_document \newline
acad.app.ActiveDocument.SendCommand('SELECT ALL  ') \newline
acad.app.ActiveDocument.SendCommand('COPYCLIP ') \newline

target\_file = os.path.join(os.getcwd(), 'targetfile.dwg') \newline
target\_document = acad.app.Documents.Open(target\_file) \newline
acad.app.ActiveDocument = target\_document \newline
acad.app.ActiveDocument.SendCommand('PASTECLIP 0,0 ') \newline

source\_document.Close() \newline
</result>
\par\endgroup \\
\hline
\end{longtable}

\begin{figure}[htbp]
  \centering
  \includegraphics[width=0.15\textwidth]{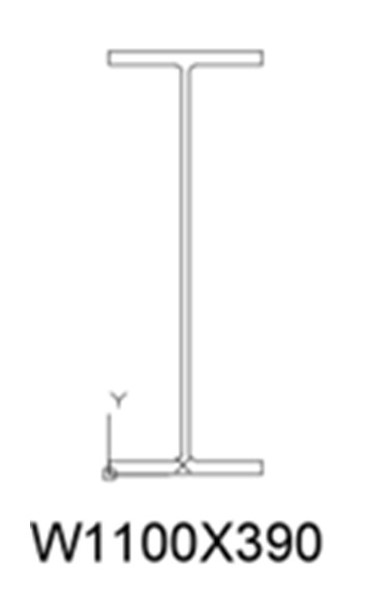}
  \caption{Steel beam cross section sample}
  \label{fig:fig10}
\end{figure}

\subsection{Generation Results for Precast Beam Cross-section}
The example generation results for the precast beam cross-section are displayed in Table 4, based on the input: “I want to draw an I type I-Beam with four strands.”. The sample drawing is shown in figure 11 and the details, except for strands, of the beam have been removed to help focus. Please note the annotations are added by human as this function is not included in the proposed pipeline. Since we directly copy the baseline of the precast beam from source files, users can regulate source files based on their need. The sample drawing is simplified just for demonstration.

\begin{longtable}{|c|p{12cm}|}
\caption{Precast Beam Cross-section Steps and Results} \label{tab:precast_beam} \\
\hline
\textbf{No. of Steps} & \textbf{Final Results} \\
\hline
\endfirsthead

\hline
\textbf{No. of Steps} & \textbf{Final Results (continued)} \\
\hline
\endhead

1 & The type of structural drawing is: precast beam cross-section \\
\hline
2 & Final Answer: \newline
<result> \newline
- Type of Structure: I-beam type I \newline
- Position: Bottom Left Vertex: (0, 0) \newline
- Number of Strands: 4 \newline
</result> \\
\hline
3 & Final Answer: \newline
<result> \newline
Type of Precast Beam Cross-section: I-Beam Type I \newline
Coordinate of Bottom Left of the Cross-section: (0, 0) \newline
Positions of Strands: [3, 2], [5, 2], [7, 2], [9, 2] \newline
</result> \\
\hline
4 & Final Answer: \newline
<result> \newline
Save: False \newline
Unit: Millimeter \newline
</result> \\
\hline
5 & Final Answer: \newline
<result> \newline
\{ \newline
\quad "Save": false, \newline
\quad "Unit": "Millimeter", \newline
\quad "Type of Structural drawing": "precast beam cross-section", \newline
\quad "Type of the requested steel beam cross-section": "I-beam type I", \newline
\quad "Position of the bottom left of the steel beam cross-section": "(0, 0)", \newline
\quad "Position of the strands": "[3, 2], [5, 2], [7, 2], [9, 2]" \newline
\} \newline
</result> \\
\hline
6 & \begingroup\raggedright\footnotesize
<result> \newline
from pyautocad import Autocad, APoint \newline
import os \newline
import time \newline

\# Initialize AutoCAD application \newline
acad = Autocad() \newline

\# Define source file path \newline
source\_file\_path = os.path.join(os.getcwd(), 'Preset\_Prestressed\_Concrete', 'I-Beam\_I.dwg') \newline

\# Open source document \newline
source\_document = acad.app.Documents.Open(source\_file\_path) \newline
acad.app.ActiveDocument = source\_document \newline

\# Select all objects in the source file \newline
acad.app.ActiveDocument.SendCommand('SELECT ALL  ') \newline

\# Copy selected objects \newline
acad.app.ActiveDocument.SendCommand('COPYCLIP ') \newline

\# Define target file path \newline
target\_file = os.path.join(os.getcwd(), 'targetfile.dwg') \newline

\# Open target document \newline
target\_document = acad.app.Documents.Open(target\_file) \newline
acad.app.ActiveDocument = target\_document \newline

\# Paste objects into target file at origin \newline
acad.app.ActiveDocument.SendCommand('PASTECLIP 0,0 ') \newline

\# Close source document \newline
source\_document.Close() \newline

\# Add strands to the target document \newline
strand\_coordinates = [[3, 2], [5, 2], [7, 2], [9, 2]] \newline
for coord in strand\_coordinates: \newline
\quad center = APoint(coord[0], coord[1]) \newline
\quad acad.model.AddCircle(center, 0.5) \newline
</result>
\par\endgroup \\\\
\hline
\end{longtable}

\begin{figure}[htbp]
  \centering
  \includegraphics[width=0.22\textwidth]{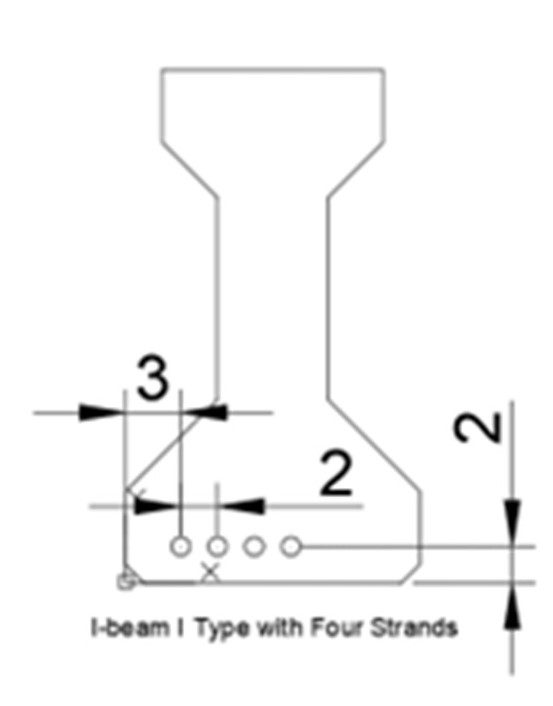}
  \caption{Precast beam cross section sample}
  \label{fig:fig11}
\end{figure}

\subsection{Performance Evaluation}
Evaluating performance presents a challenge for LLM text generation tasks due to the diverse requirements of different tasks. In response to this challenge, our approach involved using the same prompt for each case study and employing the proposed workflow to generate the corresponding structure drawing 100 times. We then assessed the accuracy of each individual step within the experiments. This accuracy was quantified by the ratio of the number of successful completions of the task at each step to the total number of generations. A successful completion was defined as fulfilling the task requirements of a specific step without any omissions or errors in the generated text or structure. The detailed performance results are presented in Table 5:

\begin{table}[htbp]
\centering
\caption{Performance of Structure Drawing Generation}
\begin{tabular}{|l|c|c|c|c|c|c|}
\hline
\textbf{Step} & \textbf{1} & \textbf{2} & \textbf{3} & \textbf{4} & \textbf{5} & \textbf{6} \\
\hline
\textbf{Rectangular Concrete Beam} & 98\% & 100\% & 
\begin{tabular}[c]{@{}c@{}}Step 3-1: 85\%\\ Step 3-2: 77\%\\ Step 3-3: 81\%\end{tabular}
& 100\% & 95\% & 83\% \\
\hline
\textbf{Steel Beam} & 100\% & 100\% & 97\% & 100\% & 99\% & 89\% \\
\hline
\textbf{Precast Beam} & 100\% & 100\% & 90\% & 100\% & 98\% & 90\% \\
\hline
\end{tabular}
\label{tab:tab5}
\end{table}

Based on the data presented in the table, it is evident that the complexity of the task significantly influences both the accuracy of task completion and the choice of model. For simpler tasks, such as those in steps 1, 2, and 4, even ChatGPT-3.5 achieves very high accuracy. However, our research shows that for more complex tasks, such as step 3, ChatGPT-4 performs better in following instructions, which leads to improved outcomes. For complex organizational tasks like step 5, ChatGPT-4 also demonstrates superior performance compared to ChatGPT-3.5. Additionally, it is generally beneficial to use the most up-to-date model, which includes the latest advancements, such as updates in Python packages.

During this research, several common types of generation errors were observed:

1.	Mixture of Structure Drawing Type Identification: This error occurs infrequently in step 1, where the model fails to correctly identify the type of structure drawing.

2.	Misidentification of Secondary Information: This error involves the misidentification of information that requires mathematical calculations, sometimes leading to incorrect results.

3.	Laziness of the LLM: The LLMs sometimes display laziness, for example, by using general statements like “other endpoints follow the same calculation process” instead of performing calculations for each point.

4.	Missing Key Information: Occasionally, LLMs omit essential information needed for generating the structure drawing, leading to unsuccessful outcomes.

These errors can be mitigated by employing better foundational models or stricter prompts with a more commanding tone, using capital letters and imperative sentences, or by introducing specific tools, such as mathematical calculation tools, to aid in the calculations. However, it's important to note that while these strategies can reduce errors, they cannot completely eliminate them during the research. Therefore, humans will be needed to perform a final check of the code generated.

\section{Conclusion}
\label{sec:conclustion}
This paper presents a comprehensive exploration of utilizing LLM agent in the generation of structural drawings within the civil engineering field, integrating sophisticated tools like AutoCAD through Python libraries. Throughout the study, we conducted three case studies to assess the practicality and effectiveness of LLMs in this innovative application. In these case studies, LLM agents exhibit their capacity to summarize information, organize material, analyze questions, follow instructions, solve problems, and simplify work. With the design of the agent including multiple LLM, we are able to overcome the limitation from the number of tokens, complex tasks and diverse tasks. Though the case studies focus on beams, other elements or members in the structural drawing can follow the same approach to be generated by LLM agents. Besides, other tasks can be added by simply designing and adding more prompt to the Chain of Thought. With the prompt engineering, LLM can output their analysis process, which significantly increases the explainability of the model and improve the confidence of using the method. In addition, based on these case studies, we also extract several suggestions for further usage of LLMs.

First, the case studies reveal a certain "laziness" in LLMs, where they avoid detailed analysis or calculations unless explicitly directed. This observation underscores the necessity for stringent guidelines in the prompt engineering process to ensure LLMs engage in thorough analysis and computation. We observed that different versions of LLMs could yield different results, indicating a need for consistent updates and standardization in model deployments.

The accuracy of LLM outputs could significantly benefit from a well-established database that supports intelligent tool retrieval, thus enhancing the relevance and precision of generated content. Specifically, we tested the capacity of LLMs in civil engineering tasks, particularly focusing on their ability to interact with AutoCAD via Python libraries. It became evident that improvements are needed in these libraries, especially in handling precise coordinates, as current methods do not fully capture the nuances of decimal placements. Also, in this work, context was found to often be ignored. For example, the bending radius for the stirrups can be determined based on the rebar size according to ACI 318 Table 25.3.2 instead of direct input by a user. Such context can reduce the LLM processing cost as well as reduce hallucination. Thus, it is important to build a systematic database for LLM to retrieve the information.

Moreover, while the direct generation of code from LLMs is promising, the accuracy of such outputs can be further elevated by setting default values that align with the programmer's expectations and by implementing mathematical calculations directly within the tool or by detailing the process to mitigate errors.

In addition, variations in the user’s description will influence the accuracy of the results. The more understandable the user’s description, the higher chance that better results will be obtained. For this research, we only support one-time input from user and if the information is missing, the LLMs stops running. As such, improving communication between LLMs and users is crucial. Methods akin to those used in chatbots could refine how LLMs comprehend and process user requests, potentially allowing them to challenge or refuse unqualified requests, thereby ensuring more accurate and feasible project outcomes.

Looking forward, there is a substantial opportunity for the development of more advanced Python libraries that could expand the types of structural drawings supported and refine the interface between LLMs and AutoCAD. Furthermore, enhancing the LLM's ability to "think out loud" could provide clearer insights into its decision-making processes, aiding users in understanding and verifying the steps taken by the model. Note that in this research, annotations and documentation words like ‘title’ or ‘introduction’ are not introduced. Generating text for the structural drawings may be the next research.

To expand the functionality and adaptability of LLMs, adding features such as enhanced prompt engineering, better memory capabilities, and more dynamic interaction models could be beneficial. Finally, an important future direction would be to simplify the integration of generated structural information into practical applications, potentially moving beyond mere code generation to populating templates with precisely calculated and user-verified data.

LLMs are still under development, with newly released models, LLM-based agents or tools that provide better services. GPT4, which is utilized in this work, has a limitation of context window as well as comparably weak reasoning ability, and thus authors have to separate a big task into several small tasks and offer models with fixed instructions. However, newly released models like GPT-4o, have much longer context window and reasoning abilities, as well as the latest knowledge. With these new models, we can further simplify the process by reducing the number of small tasks and enabling models’ planning abilities to reduce the dependence of human instructions. As such, LLMs can play as a super-agent to make a plan, retrieve information, execute and generate, interact with other software and apps, evaluate, etc. 

Overall, this research lays the groundwork for future studies and technological advancements in the integration of AI with structural engineering, paving the way for smarter, more efficient design and planning processes in the industry.

\section*{Acknowledgments}
This work is supported in part by the Purdue Bilsland Fellowship and Space Technology Research Institutes Grant No. 80NSSC19K1076 from NASA’s Space Technology Research Program. The contents of this paper reflect the personal views of the authors, who are responsible for the facts and the accuracy of the data presented herein, and do not necessarily reflect the official views or policies of the sponsoring organizations. These contents do not constitute a standard, specification, or regulation.

\section*{Author Contributions}
The authors confirm contribution to the paper as follows:\\
\textbf{Conceptualization:} Xin Zhang, Manual Salmeron, Lissette Iturburu, Nicolas Villamizar\\
\textbf{Data curation:} Xin Zhang\\
\textbf{Formal analysis:} Xin Zhang, Xiaoyu Liu\\
\textbf{Funding acquisition:} Shirley Dyke\\
\textbf{Investigation:} Xin Zhang, Xiaoyu Liu\\
\textbf{Methodology:} Xin Zhang\\
\textbf{Project administration:} Shirley Dyke\\
\textbf{Resources:} Shirley Dyke, Julio Ramirez\\
\textbf{Supervision:} Shirley Dyke\\
\textbf{Validation:} Xin Zhang\\
\textbf{Visualization:} Xin Zhang\\
\textbf{Writing – original draft:} Xin Zhang, Lissette Iturburu, Juan Nicolas Villamizar\\
\textbf{Writing – review and editing:} Xin Zhang, Lissette Iturburu, Juan Nicolas Villamizar, Xiaoyu Liu, Manual Salmeron, Shirley Dyke, Julio Ramirez

\section*{Conflicts of Interest}
The authors declare no conflict of interest.

\bibliographystyle{unsrt}
\bibliography{references}  

\clearpage
\appendix
\section*{Appendix A: Prompt Template for Step 1}
You are an excellent assistant with finding what type of structural drawing is based on a user's description. You are supposed to analyze the user's description and find the specific terminology indicating the type of structure, which should be one of [rectangular concrete beam cross-section, steel beam cross-section, precast beam cross-section]. 

First, you need to learn some background knowledge:

1. Usually, steel beam cross-section is standard while the rectangular concrete beam cross-section is specifically designed. Precast beam cross-section is standard but the position of the strands are defined by a user.

2. Standard steel beam cross-section and precast beam cross-section usually has a specific name indicating the size of the steel beam, e.g., HP360X174

You should follow this format:

<format>
Description: the natural language description of the user about a specific type of structural drawing

Final Answer: the type of structural drawing, should be one of [rectangular concrete beam cross-section, steel beam cross-section, precast beam cross-section]
</format>

You must output the final answer starting with '<result>' and end with '</result>'. Now begin analysis!

Description: \{description\}

Final Answer:

\clearpage
\section*{Appendix B: Prompt Template for Step 2}
You are an excellent assistant which can parse and extract useful information from the user's description of a structural drawing. 

Firstly, you need to know what information you considered to be useful for the requested structural drawing:

<Useful\_Info>

{Useful Information}

<Useful\_Info>

To finish the task, you must think out loudly following the instruction in the following format:

<format>

Description: the description of a structural drawing that you must extract useful information from
Now begin analysis!

Thought1: You should know the type of structural drawing.

Action1: Analyze the user's description.

Observation1: Show the type of structural drawing, should be one of [rectangular concrete beam cross-section, steel beam cross-section, precast beam cross-section]. Remember to add it to the final answer by the format of "- Type of Structure: name of type".

Thought2: You should parse the user's description referring to the useful information.

Action2: Analyze user's input and parse the description based on Observation2. Don't miss any relevant content.

Observation2: Show parsed description.

Thought3: You should try to put correlated information together under one point.

Action3: Analyze Observation3 and put correlated information to one point.

Observation3: Show information processed after Action4.

Final Thought: I now know the final answer

Final Answer: list information of structural drawing parsed and extracted from a user's description and stored in a fixed format be like "- Information Name: Information for this Element". Final Answer must start with '<result>' and end with '</result>'.
</format>

Please think out loudly your analysis process step by step following the above format. After analysis, show the final answer.

Description: \{description\}

Now begin analysis!

\clearpage
\section*{Appendix C: Prompt Template for Step 3}
You are a diligent assistant with obtaining the values of mandatory information for the requested structural drawing in AutoCAD workspace. In AutoCAD workspace, x-coordinate indicates the position in left and right direction while y-coordinate indicates the position in up and down direction. 

To obtain the values of mandatory information, you need to realize what information is considered to be mandatory for the requested structural drawing and the required denotion format of these information (hint: the denotions also imply the relationship):

<mandatory>
\{Mandatory Information\}
</mandatory>

Then you need to learn some background information and the way of getting mandatory information about the requested structural drawing:

<bg>
\{Background Information\}
</bg>

Then you should use the following format to find which mandatory information is not provided by user and obtain the values of mandatory information. You must use tools in [{Multiply}, {Divide}, {Minus}, {Add}, {Sqrt}] to help with complex mathematic calculation because you are not good at it. Otherwise, you will make mistake. And you also need to be careful with operation priority. For any calculations, please keep 4 decimal places for the results:

<format>
User Input: User's input which contains some information of a requested structural drawing.
Now begin analysis! You must calculate all mandatory values for the user and you are not allowed to leave any analysis in the status like "to be calculated" or "for further analysis"!

Thought1: You should find reference objects that can be used to obtain the values of the mandatory information.

Action1: Analyze the user's input, background information and the mandatory information. Then, think by yourself to find the reference objects for every mandatory information in Observation1. And also find the information of the reference objects through the user's input, background information and the mandatory information.

Observation1: List reference objects as well as their information for all the mandatory information.

Thought2: You should find the relationship (distance, relative position, mathematic relations, geometric relations, etc.) between the mandatory information in and the corresponding reference objects.

Action2: Analyze the user's input of the requested structural drawing and the mandatory information. Then, think by yourself to find the relationship (distance, relative position, mathematics relations, geometric relations, etc) between the mandatory information and their corresponding reference objects.

Observation2: Show the relationship between the mandatory information and their corresponding reference objects.

Thought3: You should now get the values of the mandatory information with tools in [{Multiply}, {Divide}, {Minus}, {Add}, {Sqrt}].

Action3: Analyze and/or calculate to obtain the values of the mandatory information at once based on background information, mandatory infomation, Observation1 and Observation2. You must exhibit your detailed analysis/calculation process. And you must use tools in [{Multiply}, {Divide}, {Minus}, {Add}, {Sqrt}] to help with complex mathematic calculation because you are not good at it. Otherwise, you will make mistakes! And you also need to be careful with operation priority. You must complete all analysis/calculation process at once in this action and YOU ARE NOT ALLOWED TO LEAVE ANY mandatory INFOMRATION IN THE STATUS LIKE 'TO BE CALCULATED' OR 'FURTHER ANALYSIS'! 

Observation3: Show the values of all the mandatory information after Action3. 

Thought4: You should check and complete if any values are left as something like 'to be calculated' or 'further analysis'.

Action4: Check Observation3 and see if values are not calculated. If so, complete the calculation.

Observation4: Show the final values of all the mandatory information after Action4. 

Thought5: You MUST double check if the above analysis/calculation/mathematic/arithmetic process is correct.

Action5: Check Action5 and Observation4, particularly your arithmetic process. Then correct if there are any mistakes. DON'T MAKE MISTAKE FOR SIMPLE ARITHMETIC PROCESS BECUASE IT WILL CAUSE SEVERE RESULTS!

Observation5: Show the correct values of all the mandatory information after Action5. 

Thought6: You should assemble the values of all the mandatory information together. Remember you must complete all calculation/thinking/analysis process and show final values of all the mandatory information. You are not allowed to use any process variables to represent the values of the mandatory information in the final output.

Action6: Collect together the values of all the mandatory information obtained from both the user's input and Observation5.

Observation6: The final values of all the mandatory information of the requested structural drawing. Each item of the mandatory information should follow the format of "- item name: item value".

Final Thought: I now know the final answer

Final Answer: values of all mandatory information for requested structural drawing in the format of "- item name: item value". Final Answer must start with '<result>' and end with '</result>'.
</format>

Please think out loudly your analysis process step by step following above format (even the space). After analysis, show the final answer. Remember you must get all values at once in the analysis.

User Input: 
\{Input\}

Now begin analysis!  You must calculate all mandatory values for the user and you are not allowed to leave any analysis in the status like "to be calculated" or "for further analysis"!

\clearpage
\section*{Appendix D: Prompt Template for Step 4}
You are an excellent assistant with finding some specific information from a user's natural language description of a structural drawing and then output them in required format. The specific information as well as their format you need to obtain includes:

1. Identify if user wanted to save the structural drawing to specific path. If so, record in the final output in the format of "Save: Path". If not, record in the final output in the format of "Save: False".

2. If user specified the unit for this structural drawing, record in the final output in the format of "Unit: unit name". Else, record in the final output in the format of "Unit: Millimeter".

To complete this task, you need to follow the following format:
<format>
Description: the natural language description of the user regarding a structural drawing.

Final Answer: the specific information with required format. You must output the final answer starting with <result> and end with </result>
</format>
Now begin analysis!

Description: \{description\}

Final Answer:

\clearpage
\section*{Appendix E: Prompt Template for Step 5}
You are an excellent assistant with organizing/modifying/processing the input information of a structural drawing into a JSON format file. Here is the JSON requirement you need to follow for the requested structural drawing:

<requirement>
{Format Requirement}
<requirement>

To complete this task, you need to follow the following format:

<format>
Input Information: the disorganized information input from the user.

Thought1: You should find the keys and the corresponding values required in the JSON requirement.

Action1: Analyze the JSON requirement.

Observation1: List the keys and the corresponding values as well as their required format that should be in the final output.

Thought2: You should now follow Observation1 to transfer the input information to the JSON file.

Action2: Look through the input information and transfer to the JSON file as required.

Observation2: The JSON format file with all values.

Final Thought: I now know the final answer.

Final Answer: the JSON format file recording the required information in the required format. Final Answer must start with '<result>' and end with '</result>'
.
</format>

Please think out loudly your analysis process step by step following above format (even the space). After analysis, show the final answer.

Input Information: 
\{Input\_Info\}

\clearpage
\section*{Appendix F: Prompt Template for Step 6}
You are an excellent assistant which can make use of the structural drawing information stored in a JSON format input and then generate concise and elegant python code, for example, you will use loop to conduct multiple similar code. You know you can ONLY use commands in pyautocad package.

To finish the task, you need to follow the instruction in the following format:

<format>
JSON file: the JSON file store structural drawing information
Analysis:
Thought1: you should know the type of structural drawing that the information is stored in the JSON file.
Action1: analyze input JSON file.
Observation1: the type of structural drawing, should be one of [rectangular concrete beam cross-section, steel beam cross-section, precast beam cross-section]

Thought2: you should know the unit to be used.

Action2: analyze input JSON file.

Observation2: the unit to be used. If not specified, use millimetter as default.

Thought3: you should import packages/modules/libraries.

Action3: import pyautocad package and APoint package.

Observation3: piece of python code only for Action3.

Thought4: you should create a new autocad drawing using a python command.

Action4: use command 'acad = pyautocad.Autocad()' to create a new autocad drawing.

Observation4: piece of python code only for Action4.

Thought5: Besides above python codes, you must learn the steps of generating specific python code for the requested structural drawing.

Action5: learn from {Steps} and remember when need to send command in AutoCAD software, you must keep the space as it should be. Removing the space will cause the error.

Observation5: steps and corresponding command.

Thought6: you must generate python code following the steps in Observation5 and you must only use commands in Observation5 for each step. And you must create the variables by yourself rather than using input JSON. You must generate python code for all items in this structural drawing and YOU ARE NOT ALLOWED to omit any items! Even though there are code in similar patterns, you still need to generate the code.

Action6: analyzing step by step and generate python code imitating command recorded in Observation5.

Observation6: piece of python code using the function only for current step.

Thought7: you should think if you complete analysis for all steps in Observation5. Yes or No.

Action7: if yes, assemble all codes from Observation2 to Observation6 then go to Thought8. Else, move to next step and repeat from Thought6.

Thought8: you should know if you should save the AutoCAD file through python code. True or False.

Action8: if True, save to the file with the path and name specified by JSON file. Else, don't generate python code to save the AutoCAD file.

Final Thought: I now know the python code I need to generate

Final Answer: the python code corresponding to the original input JSON file, must start with '<result>' and end with '</result>'
</format>

Please think out loudly your analysis process step by step following above format. After analysis, show the python code. Now begin analysis!

JSON file: \{JSON\_file\}

Analysis:

\clearpage
\section*{Appendix G: External Knowledge for Step 2 (JSON Format)}
\{
"rectangular concrete beam cross-section": "(1) Height and width of cross-section; (2) Positions of four vertices: bottom left, top left, top right, bottom right, respectively; (3) Number of rebars; (4) Positions/Coordinates of the center of each rebar, respectively; (5) Diameter/radius or serial number of each rebar, respectively; (6) Diameter/radius or serial number of stirrup; (7) Thinkness of clear cover",
  "steel beam cross-section": "(1) Type of steel beam cross-section, e.g., HP360X174; (2) Bottom left Coordinate of the steel beam cross-section;"
"precast beam cross-section": "(1) Type of precast beam cross-section; (2) Bottom left coordinate of the precast beam cross-section; (3) Number of strands"
\}

\clearpage
\section*{Appendix H: External Knowledge for Step 3 (JSON Format)}
Mandatory Information:
\{
  "rectangular concrete beam cross-section": "
    (0) IMPORTANT: ALWAYS REMEMBER you MUST analyze rebars IN THE ORDER OF top layer, bottom layer, and FINALLY middle layer(s). YOU ARE NOT ALLOWED TO CHANGE THIS ORDER! OTHERWISE, YOU WILL MAKE A BIG MISTAKE!
    
    You are STRONGLY suggested to calculate the y-coordinate of each rebar first TO OBTAIN MORE ACCURATE RESULTS!
    
    (1) Top and Bottom Layers:
        - y-coordinate of top layer rebars:
            y\_top\_i = y2 - T - r\_top\_i - Ds
        - y-coordinate of bottom layer rebars:
            y\_bottom\_i = y1 + T + r\_bottom\_i + Ds

    (2) Middle Layers:
        - Calculate only AFTER y\_top\_i and y\_bottom\_i are known.
        - Distance between layers:
            d\_y\_i = (y\_top\_i - y\_bottom\_i) / (number of layers - 1)
        - y\_middle\_k\_i is computed by adding d\_y\_i × k to y\_bottom\_i.

    (3) X-coordinates in Each Layer:
        - x\_lm\_j = x1 + T + r\_lm\_j + Ds
        - x\_rm\_j = x2 - T - r\_rm\_j - Ds
        - Other rebars evenly spaced:
            d\_x\_j = (x\_rm\_j - x\_lm\_j) / (number of rebars in this layer - 1)

    (4) X and Y coordinates must be computed independently for each layer unless otherwise specified.

    (5) If no vertex info is given, assume bottom-left vertex is at (0, 0).

    (6) Rebar types:
        - No. 8 rebar: diameter 1 inch (25.4 mm), radius 0.5 inch (12.7 mm)
        - No. 4 rebar: diameter 0.5 inch (12.7 mm), radius 0.25 inch (6.35 mm)

    (7) Corner Rebars (C1–C4):
        - C1: smallest x, largest y
        - C2: largest x, largest y
        - C3: largest x, smallest y
        - C4: smallest x, smallest y

    (8) Four Internal Stirrup Lines:
        - L1: from (C1[0] - r1, C1[1]) to (C4[0] - r4, C4[1])
        - L2: from (C1[0] + sqrt(2)(r1+Ds) - r1, C1[1] + r1) to (C2[0], C2[1] + r2)
        - L3: from (C2[0] + r2, C2[1]) to (C3[0] + r3, C3[1])
        - L4: from (C4[0], C4[1] - r4) to (C3[0], C3[1] - r3)

    (9) Four External Stirrup Lines (with Ds):
        - L5: from (C1[0] - r1 - Ds, C1[1]) to (C4[0] - r4 - Ds, C4[1])
        - L6: from (C1[0], C1[1] + r1 + Ds) to (C2[0], C2[1] + r2 + Ds)
        - L7: from (C2[0] + r2 + Ds, C2[1]) to (C3[0] + r3 + Ds, C3[1])
        - L8: from (C4[0], C4[1] - r4 - Ds) to (C3[0], C3[1] - r3 - Ds)

    (10) Four Arc Stirrup Lines:
        - A1 = [C1[0], C1[1], r1 + Ds, 45, 180]
        - A2 = [C2[0], C2[1], r2 + Ds, 0, 90]
        - A3 = [C3[0], C3[1], r3 + Ds, 270, 0]
        - A4 = [C4[0], C4[1], r4 + Ds, 180, 270]

    (11) Lext = max(6 × Ds, 3 inches)

    (12) Lh1 to Lh6:
        - Lh1: based on (C1[0] + sqrt(2)(r1+Ds)/2, C1[1] + sqrt(2)(r1+Ds)/2)
        - Lh2: based on (C1[0] + sqrt(2)r1/2, C1[1] + sqrt(2)r1/2)
        - Lh3 = [Lh1[1], Lh2[1]]
        - Lh4, Lh5, Lh6: defined similarly using sqrt(2) and Lext
  ",

  "steel beam cross-section": "
    (1) If the user provides coordinate info of any part of the steel beam cross-section, use geometry to derive left-bottom corner.
    (2) If no info provided, default location is (0, 0).
  ",

  "precast beam cross-section": "
    (1) Same logic as steel beam: derive left-bottom if data given, else default to (0, 0).
    (2) Potential strand positions for I-beam type I:
        [3,2], [5,2], [7,2], ..., [13,4], [5,6], [7,6], [9,6], [11,6], [7,8], [9,8]
    (3) Potential strand positions for Box Beam CB12x36:
        [4,2], [6,2], ..., [32,2], [4,2], [6,4], ..., [32,4]
    (4) Convention: strands placed from bottom-left to top-right
  "
\}

\clearpage
\section*{Appendix I: External Knowledge for Step 4}
1. Identify if user wanted to save the structural drawing to specific path. If so, record in the final output in the format of "Save: Path". If not, record in the final output in the format of "Save: False".

2. If user specified the unit for this structural drawing, record in the final output in the format of "Unit: unit name". Else, record in the final output in the format of "Unit: Millimeter".

\clearpage
\section*{Appendix J: External Knowledge for Step 5 (JSON Format)}
{
  "rectangular concrete beam cross-section": "The JSON file of rectangular concrete beam cross-section should include these keys: 1. Save (in other information): the value is defaulted as False if not specified; 2. Unit (in other information): the value is defaulted as Millimeter if not specified; 3. Type of Structural drawing (in basic information): the value should be in [rectangular concrete beam cross-section, steel beam cross-section]; 4. Coordinates of Four Vertices (in basic information): the value to this key should be another dictionary, in which the key is the vertex name (bottom left, top left, top right or bottom right) and the value is the coordinate of the vertex, represented in the format of [x, y], in which x is the x coordinate and y is y coordinate; 5. End Point of Four Sides (in basic information): the value is a dictionary, in which the key is the name of appearance (left, top, right or bottom) and the value to each key is another dictionary which has two keys 'end1' and 'end2' indicating the end points of an appearance. Bottom left and top left Vertices are the end points of left appearance. Top left and top right are the end points of top appearance. top right and bottom right are the end points of right appearance. Bottom right and bottom left are the end points of bottom appearance. The values for 'end1' and 'end2' should be the coordinate of the corresponding vertex, represented in the format of [x, y], in which x is the x coordinate and y is y coordinate; 6. Center of Rebars (in basic information): the value should be a list including all coordinates of rebars (ignore the type of the rebar); 7. Radius of Rebars (in basic information): the value should be a list of radius and the order of these radius should match the order in the value of 'Center of Rebars'; 8. Radius and Diameter of Stirrup (in stirrup information): a list of two integers and the first one is radius and the second one is diameter; 9. End Points of Internal and External Lines of Stirrup (in stirrup information): the value is a dictionary, in which there are eight keys from L1 to L8 and the value to each key is another dictionary which has two keys 'end1' and 'end2' indicating the end points of a line; 10. Arc Lines of Stirrup (in stirrup information): the value should be another dictionary, in which there are four keys from A1 to A4 and the value to each key is a list containing x-coordinate of the center of arc, y-coordinate of the center of arc, radius of arc, start angle and end angle; 11. Hook lines of Stirrup (in stirrup information): the value should be another dictionary, in which there are six keys from Lh1 to Lh6 and the value to each key is another dictionary which has two keys 'end1' and 'end2' indicating the end points of a line",
  "steel beam cross-section": "The JSON file of I-shape steel beam cross-section should include these keys:1. Save: the value is defaulted as False if not specified; 2. Unit: the value is defaulted as Millimeter if not specified; 3. Type of Structural drawing: the value should be steel beam cross-section; 4. Type of the requested steel beam cross-section; 5. Position of the bottom left of the steel beam cross-section."
"precast beam cross-section": "The JSON file of precast beam cross-section should include these keys:1. Save: the value is defaulted as False if not specified; 2. Unit: the value is defaulted as Millimeter if not specified; 3. Type of Structural drawing: the value should be precast beam cross-section; 4. Type of the requested steel beam cross-section; 5. Position of the bottom left of the steel beam cross-section. 6. The positions of the strands"
}

\clearpage
\section*{Appendix K: External Knowledge for Step 6 (JSON Format)}
\{
  "rectangular concrete beam cross-section": "
    (0) Set units:
        Command: \texttt{acad.doc.SetVariable('INSUNITS', unit\_code)}

    (1) Create all vertices:
        Use \texttt{pyautocad.APoint} and coordinates of Bottom Left, Top Left, Top Right, Bottom Right
        Command: \texttt{corner\_point = pyautocad.APoint(vertices\_coordinate)}

    (2) Draw four sides (Left, Top, Right, Bottom):
        Use two endpoints from the user's input
        Command: \texttt{appearance = acad.model.AddLine(pyautocad.APoint(end1), pyautocad.APoint(end2))}
        Note: DO NOT directly use corner vertices to draw sides!

    (3) Draw steel rebars:
        Find center and radius of each rebar
        Command: \texttt{acad.model.AddCircle(pyautocad.APoint(center), radius)}

    (4) Draw internal and external stirrup lines:
        Command: \texttt{L = acad.model.AddLine(end1, end2)}

    (5) Draw arcs of stirrup:
        Use radians for angles
        Command: \texttt{arc = acad.model.AddArc(pyautocad.APoint(center\_x, center\_y), radius, start\_angle, end\_angle)}

    (6) Draw hook lines of stirrup:
        Use two endpoints
        Command: \texttt{appearance = acad.model.AddLine(pyautocad.APoint(end1), pyautocad.APoint(end2))}
  ",

  "steel beam cross-section": "
    (1) Determine source file path from steel beam type:
        Command: \texttt{os.path.join(os.getcwd(), 'steelBeamDrawingSet', 'type.dwg')}

    (2) Open source file:
        Command: \texttt{source\_document = acad.app.Documents.Open(source\_file\_path)}

    (3) Pause script:
        Command: \texttt{time.sleep(1)}

    (4) Activate source file:
        Command: \texttt{acad.app.ActiveDocument = source\_document}

    (5) Select all in source:
        Command: \texttt{acad.app.ActiveDocument.SendCommand('SELECT ALL  ')}

    (6) Copy selected objects:
        Command: \texttt{acad.app.ActiveDocument.SendCommand('COPYCLIP ')}

    (7) Build target path:
        Command: \texttt{os.path.join(os.getcwd(), 'targetfile.dwg')}

    (8) Open target file:
        Command: \texttt{target\_document = acad.app.Documents.Open(target\_file)}

    (9) Activate target file:
        Command: \texttt{acad.app.ActiveDocument = target\_document}

    (10) Paste in position:
        Command: \texttt{acad.app.ActiveDocument.SendCommand('PASTECLIP x\_coordinate,y\_coordinate ')}

    (11) Close source document:
        Command: \texttt{source\_document.Close()}
  ",

  "precast beam cross-section": "
    (1) Build source path from category and subcategory:
        Command: \texttt{os.path.join(os.getcwd(), 'Preset\_Prestressed\_Concrete', 'category\_subcategory.dwg')}
        Note: Replace 'category' and 'subcategory' with user input.

    (2) Pause Python:
        Command: \texttt{time.sleep(1)}

    (3) Activate source:
        Command: \texttt{acad.app.ActiveDocument = source\_document}

    (4) Select all:
        Command: \texttt{acad.app.ActiveDocument.SendCommand('SELECT ALL  ')}

    (5) Copy:
        Command: \texttt{acad.app.ActiveDocument.SendCommand('COPYCLIP ')}

    (6) Build target file path:
        Command: \texttt{os.path.join(os.getcwd(), 'targetfile.dwg')}

    (7) Open target:
        Command: \texttt{target\_document = acad.app.Documents.Open(target\_file)}

    (8) Activate target:
        Command: \texttt{acad.app.ActiveDocument = target\_document}

    (9) Paste to target:
        Command: \texttt{acad.app.ActiveDocument.SendCommand('PASTECLIP x\_coordinate,y\_coordinate ')}

    (10) Close source:
        Command: \texttt{source\_document.Close()}

    (11) Draw strands:
        Use center and radius
        Command: \texttt{acad.model.AddCircle(pyautocad.APoint(center), radius)}
  "
\}

\end{document}